\documentclass[conference]{IEEEtran}
\usepackage{times}

\usepackage[numbers]{natbib}
\usepackage{multicol}
\usepackage[pagebackref=true,breaklinks=true,colorlinks,bookmarks=false]{hyperref}
\usepackage{amsthm}
\usepackage{adjustbox}
\usepackage{caption}
\usepackage{stackengine}
\usepackage{makecell}
\usepackage{caption}
\usepackage{comment}
\usepackage{siunitx}
\usepackage{relsize}
\usepackage{ifthen}
\usepackage[colorinlistoftodos]{todonotes}

\usepackage[vlined,ruled,linesnumbered]{algorithm2e}
\usepackage{graphics} % for pdf, bitmapped graphics files
\usepackage{rotating}
\usepackage{color}
\usepackage{enumerate}
\usepackage[T1]{fontenc}
\usepackage{psfrag}
\usepackage{epsfig} % for postscript graphics files
\usepackage{booktabs}
\usepackage{graphicx,url}
\usepackage{multirow}
\usepackage{array}
\usepackage{latexsym}
\usepackage{amsfonts}
\usepackage{amsmath}
\usepackage{amssymb}
\usepackage{mathtools}
\usepackage{xstring}
\usepackage[noend]{algorithmic}
\usepackage{multirow}
\usepackage{xcolor}
\usepackage{prettyref}
\usepackage{flexisym}
\usepackage{bigdelim}
\usepackage{breqn} % load this last
\usepackage{listings}
\let\labelindent\relax
\usepackage{enumitem}
\usepackage{xspace}
\usepackage{bm}
\graphicspath{{./figures/}}
\usepackage{tikz}
\usetikzlibrary{matrix,calc}
\usepackage{tabularx}
\usepackage[scr=boondox]{mathalfa}
\usepackage{stackengine}

\usepackage{mdwlist}

\let\itemize\undefined
\makecompactlist{itemize}{stditemize}

\newrefformat{prob}{Problem\,\ref{#1}}
\newrefformat{def}{Definition\,\ref{#1}}
\newrefformat{sec}{Section\,\ref{#1}}
\newrefformat{sub}{Section\,\ref{#1}}
\newrefformat{prop}{Proposition\,\ref{#1}}
\newrefformat{app}{Appendix\,\ref{#1}}
\newrefformat{alg}{Algorithm\,\ref{#1}}
\newrefformat{cor}{Corollary\,\ref{#1}}
\newrefformat{thm}{Theorem\,\ref{#1}}
\newrefformat{lem}{Lemma\,\ref{#1}}
\newrefformat{fig}{Fig.\,\ref{#1}}
\newrefformat{tab}{Table\,\ref{#1}}

\newtheorem{theorem}{Theorem}

\newtheorem{definition}[theorem]{Definition}
\newtheorem{proposition}[theorem]{Proposition}

\newtheorem{example}[theorem]{Example}

\newcommand{\cf}{\emph{cf.}\xspace}

\newcommand{\bdmath}{\begin{dmath}}
\newcommand{\edmath}{\end{dmath}}
\newcommand{\beq}{\begin{equation}}
\newcommand{\eeq}{\end{equation}}
\newcommand{\bdm}{\begin{displaymath}}
\newcommand{\edm}{\end{displaymath}}
\newcommand{\bea}{\begin{eqnarray}}
\newcommand{\eea}{\end{eqnarray}}
\newcommand{\beal}{\beq \begin{array}{ll}}
\newcommand{\eeal}{\end{array} \eeq}
\newcommand{\beas}{\begin{eqnarray*}}
\newcommand{\eeas}{\end{eqnarray*}}
\newcommand{\ba}{\begin{array}}
\newcommand{\ea}{\end{array}}
\newcommand{\bit}{\begin{itemize}}
\newcommand{\eit}{\end{itemize}}
\newcommand{\ben}{\begin{enumerate}}
\newcommand{\een}{\end{enumerate}}

\newcommand{\calA}{{\cal A}}

\newcommand{\calC}{{\cal C}}

\newcommand{\calG}{{\cal G}}
\newcommand{\calH}{{\cal H}}

\newcommand{\ie}{\emph{i.e.,}\xspace}

\renewcommand{\boldsymbol}[1]{{\bm #1}}
\newcommand{\hide}[1]{}

\newcommand{\hiddenText}{{\color{gray} hidden text.}}
\newcommand{\hideWithText}[1]{\hiddenText}

\newcommand{\subject}{\text{ subject to }}

\newcommand{\tran}{^{\mathsf{T}}}

\newcommand{\Real}[1]{ { {\mathbb R}^{#1} } }

\newcommand{\vy}{\boldsymbol{y}}

\newcommand{\scenario}[1]{{\smaller \sf#1}\xspace}

\newcommand{\MOSEK}{\scenario{MOSEK}}

\newcommand{\linkToPdf}[1]{\href{#1}{\ksc{(pdf)}}}
\newcommand{\linkToPpt}[1]{\href{#1}{\ksc{(ppt)}}}
\newcommand{\linkToCode}[1]{\href{#1}{\ksc{(code)}}}
\newcommand{\linkToWeb}[1]{\href{#1}{\ksc{(web)}}}
\newcommand{\linkToVideo}[1]{\href{#1}{\ksc{(video)}}}
\newcommand{\linkToMedia}[1]{\href{#1}{\ksc{(media)}}}
\newcommand{\award}[1]{\xspace} % {{\red{#1}}} % omit awards

\newcommand{\pos}{\boldsymbol{p}}
\newcommand{\R}{\mathbb{R}}

\newcommand{\mymid}{\ \middle\vert\ }
\newcommand{\cbrace}[1]{\left\{#1\right\}}

\newcommand{\bmat}{\left[ \begin{array}}
\newcommand{\emat}{\end{array}\right]}

\newcommand{\spot}{\textsc{Spot}\xspace}
\newcommand{\tssos}{\textsc{Tssos}\xspace}

\newcommand{\abs}[1]{\left|#1\right|}

\newcommand{\bbN}{\mathbb{N}}

\newcommand{\ineq}{\text{ineq}}
\newcommand{\eq}{\text{eq}}

\newcommand{\ceil}[1]{\left\lceil #1 \right\rceil}

\newcommand{\csp}{\text{csp}}
\newcommand{\nchoosek}[2]{(\substack{#1 \\ #2})}

\newcommand{\mcar}{m_c}
\newcommand{\mpole}{m_p}
\newcommand{\len}{\ell}
\newcommand{\dt}{\Delta t}
\DeclareDocumentCommand{\pos}{o}{
  \IfNoValueTF{#1}{a}{a_{#1}}
}
\DeclareDocumentCommand{\rc}{o}{
  \IfNoValueTF{#1}{r_c}{r_{c,#1}}
}
\DeclareDocumentCommand{\rs}{o}{
  \IfNoValueTF{#1}{r_s}{r_{s,#1}}
}
\DeclareDocumentCommand{\fc}{o}{
  \IfNoValueTF{#1}{f_c}{f_{c,#1}}
}
\DeclareDocumentCommand{\fs}{o}{
  \IfNoValueTF{#1}{f_s}{f_{s,#1}}
}
\DeclareDocumentCommand{\lamone}{o}{
  \IfNoValueTF{#1}{\lambda_{1}}{\lambda_{1,#1}}
}
\DeclareDocumentCommand{\lamtwo}{o}{
  \IfNoValueTF{#1}{\lambda_{2}}{\lambda_{2,#1}}
}

\DeclareDocumentCommand{\Fx}{o}{
  \IfNoValueTF{#1}{F_x}{F_{x,#1}}
}
\DeclareDocumentCommand{\Fy}{o}{
  \IfNoValueTF{#1}{F_y}{F_{y,#1}}
}
\DeclareDocumentCommand{\sx}{o}{
  \IfNoValueTF{#1}{s_x}{s_{x,#1}}
}
\DeclareDocumentCommand{\sy}{o}{
  \IfNoValueTF{#1}{s_y}{s_{y,#1}}
}
\DeclareDocumentCommand{\px}{o}{
  \IfNoValueTF{#1}{p_x}{p_{x,#1}}
}
\DeclareDocumentCommand{\py}{o}{
  \IfNoValueTF{#1}{p_y}{p_{y,#1}}
}
\DeclareDocumentCommand{\lam}{m o}{
  \IfNoValueTF{#2}{\lambda_{#1}}{\lambda_{#1,#2}}
}
\DeclareDocumentCommand{\gam}{m o}{
  \IfNoValueTF{#2}{\gamma_{#1}}{\gamma_{#1,#2}}
}

\DeclareDocumentCommand{\xr}{o}{
  \IfNoValueTF{#1}{x_r}{x_{r,#1}}
}
\DeclareDocumentCommand{\yr}{o}{
  \IfNoValueTF{#1}{y_r}{y_{r,#1}}
}
\DeclareDocumentCommand{\xl}{o}{
  \IfNoValueTF{#1}{x_l}{x_{l,#1}}
}
\DeclareDocumentCommand{\yl}{o}{
  \IfNoValueTF{#1}{y_l}{y_{l,#1}}
}
\DeclareDocumentCommand{\vxr}{o}{
  \IfNoValueTF{#1}{v_{x,r}}{v_{x,r,#1}}
}
\DeclareDocumentCommand{\vyr}{o}{
  \IfNoValueTF{#1}{v_{y,r}}{v_{y,r,#1}}
}
\DeclareDocumentCommand{\vrelr}{o}{
  \IfNoValueTF{#1}{v_{rel,r}}{v_{rel,r,#1}}
}
\DeclareDocumentCommand{\dr}{o}{
  \IfNoValueTF{#1}{d_r}{d_{r,#1}}
}
\DeclareDocumentCommand{\vx}{o}{
  \IfNoValueTF{#1}{v_x}{v_{x,#1}}
}
\DeclareDocumentCommand{\vy}{o}{
  \IfNoValueTF{#1}{v_y}{v_{y,#1}}
}
\DeclareDocumentCommand{\x}{o}{
  \IfNoValueTF{#1}{x}{x_{#1}}
}
\DeclareDocumentCommand{\y}{o}{
  \IfNoValueTF{#1}{y}{y_{#1}}
}

\newcommand{\enum}[1]{\left[ #1 \right]}
\newcommand{\init}{\mathrm{init}}

\newcommand{\loss}{\text{loss}}
\newcommand{\kkt}{\text{kkt}}

\newcommand{\mvx}{\mathbf{x}}
\newcommand{\mvy}{\mathbf{y}}
\newcommand{\mva}{\bm{\alpha}}

\newcommand{\sqbk}[1]{\left[ #1 \right]}
\newcommand{\mbS}{\mathbf{S}}

\newcommand{\ksc}[1]{{#1}}

\begin{document}

\title{\vspace{-10mm} Global Contact-Rich Planning with \\ Sparsity-Rich Semidefinite Relaxations \vspace{-2mm}}

% You will get a Paper-ID when submitting a pdf file to the conference system
% \author{Author Names Omitted for Anonymous Review. Paper-ID [353]}

\author{\authorblockN{Shucheng Kang$^{* 1}$, Guorui Liu$^{* 2}$, Heng Yang$^{1}$} 
${}^{1}$School of Engineering and Applied Sciences, Harvard University\\
% ${}^{1}$Harvard University,
${}^{2}$School of Industrial and Systems Engineering, Georgia Institute of Technology\\
${}^{*}$Equal contribution\\[1mm]
% ${}^{2}$University of Science and Technology of China\\
\texttt{\url{https://computationalrobotics.seas.harvard.edu/project-spot/}}
}

%\author{\authorblockN{Michael Shell}
%\authorblockA{School of Electrical and\\Computer Engineering\\
%Georgia Institute of Technology\\
%Atlanta, Georgia 30332--0250\\
%Email: mshell@ece.gatech.edu}
%\and
%\authorblockA{Twentieth Century Fox\\
%\and
%\authorblockN{James Kirk\\ and Montgomery Scott}
%\authorblockA{Starfleet Academy\\
%San Francisco, California 96678-2391\\

% avoiding spaces at the end of the author lines is not a problem with
% conference papers because we don't use \thanks or \IEEEmembership

% \author{\authorblockN{Shucheng Kang\authorrefmark{1},
% Xiaoyang Xu\authorrefmark{2},
% Jay Sarva\authorrefmark{3},
% Ling Liang\authorrefmark{4},
% and 
% Heng Yang\authorrefmark{1}}
% \authorblockA{\authorrefmark{1}Harvard University}
% \authorblockA{\authorrefmark{2}University of California at Santa Barbara}
% \authorblockA{\authorrefmark{3}Brown University}
% \authorblockA{\authorrefmark{4}University of Maryland at College Park}
% }
% \newcommand\blfootnote[1]{%
%   \begingroup
%   \renewcommand\thefootnote{}\footnote{#1}%
%   \addtocounter{footnote}{-1}%
%   \endgroup
% }

\newcommand\blfootnote[1]{%
  \begingroup
  \renewcommand\thefootnote{}\footnote{#1}%
  \addtocounter{footnote}{-1}%
  \endgroup
}

% \maketitle

\twocolumn[{%
\renewcommand\twocolumn[1][]{#1}%
\maketitle
% \vspace{-mm}
% \vspace{-8mm}
%!TEX root = main.tex

\begin{minipage}{\textwidth}
\vspace{-8mm}
\includegraphics[width=\linewidth]{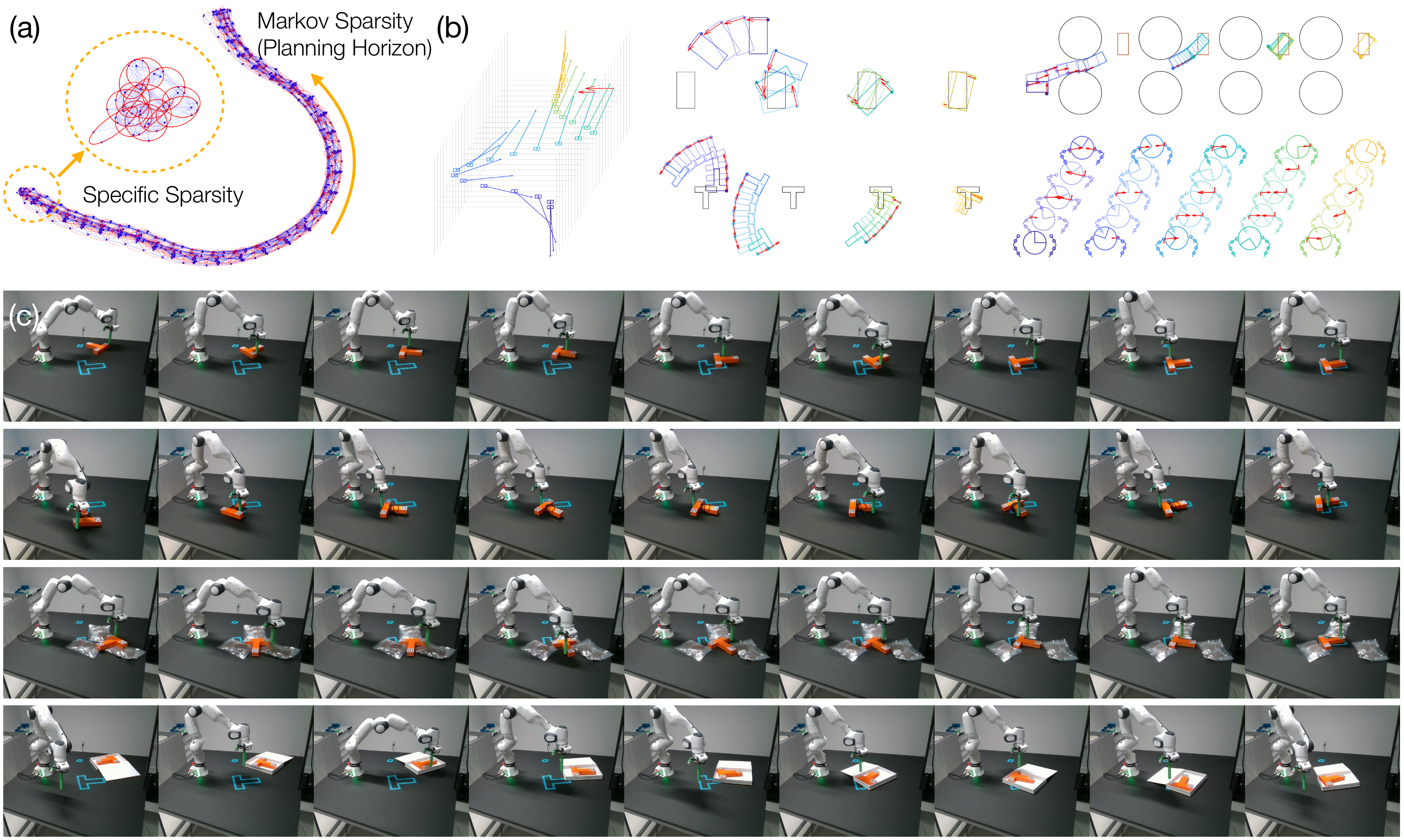}
\vspace{-6mm}
\captionof{figure}{Contact-rich planning is \emph{sparsity-rich}. (a) Sparsity graph of planar hand manipulation showing two types of sparsity. (b) Sparsity enables high-order and tight, yet small-scale, semidefinite programming (SDP) relaxations solvable by off-the-shelf SDP solvers, computing certified near globally optimal trajectories for a suite of simulated problems. (c) Real-world validation on push-T. Planning powered by global optimization succeeds the task even under severe model mismatches and disturbances (from top to bottom: clean T; T tied up by a cable; T on top of a cluttered table; and T inside a ``sliding'' box). 
\label{fig:demos}}
\vspace{-1mm}
\end{minipage}
}]

\IEEEpeerreviewmaketitle

% \blfootnote{$^*$Equal contribution}

% \input{sections/outline.tex}
%!TEX root = ../main.tex

\begin{abstract}
    We show that contact-rich motion planning is also \emph{sparsity-rich} when viewed as polynomial optimization (POP). We can exploit not only the \emph{correlative} and \emph{term} sparsity patterns that are general to all POPs, but also specialized sparsity patterns from the robot kinematic structure and the separability of contact modes. Such sparsity enables the design of high-order but sparse semidefinite programming (SDPs) relaxations---building upon Lasserre's moment and sums of squares hierarchy---that (\emph{i}) can be solved in seconds by off-the-shelf SDP solvers, and (\emph{ii}) compute near globally optimal solutions to the nonconvex contact-rich planning problems with small certified suboptimality. Through extensive experiments both in simulation (Push Bot, Push Box, Push Box with Obstacles, and Planar Hand) and real world (Push T), we demonstrate the power of using convex SDP relaxations to generate global contact-rich motion plans.
    As a contribution of independent interest, we release the Sparse Polynomial Optimization Toolbox (\spot)---implemented in C++ with interfaces to both Python and Matlab---that automates sparsity exploitation for robotics and beyond.

    % We propose a novel sparsity-rich semidefinite relaxation framework to globally and efficiently solve general contact-rich planning problems modeled by polynomial optimization problems. We extensively demonstrate that contact-rich problems are also sparsity-rich by: (1) exploring automatic correlative sparsity and term sparsity pattern generation explicitly; and (2) exploiting multi-dimensional robotics-specific sparsity patterns, such as kinematic chains and separable contact modes. The resulting small-to-medium scale SDPs can be solved in seconds while maintaining decent tightness. We further develop a novel minimizer extraction routine for correlative sparsity patterns, enabling robust extraction even with multiple global minimizers or untight relaxations. Consequently, a new C++ sparse Moment-SOS Hierarchy conversion package, Sparse Polynomial Optimization Tools (SPOT), is developed. Extensive case studies are conducted (Simulation: Push Bot, Push Box, Push Box with Obstacles, and Planar Hand; Real-world: Push T-Block), demonstrating both the tightness and efficacy of our framework.
\end{abstract}
% \vspace{-8mm}
%!TEX root = ../main.tex

\vspace{-2mm}
\section{Introduction}
\label{sec:introduction}
% \vspace{-2mm}

Contact-rich planning plays a fundamental role in robotics tasks ranging from manipulation to locomotion~\cite{mason1986ijrr-mechanics-planning-pushing,di2018iros-dynamic-locomotion,hirai1998icra-development-honda-humanoid}. At the heart of such planning problems lie two interrelated challenges: (a) \textit{Contact mode selection}: determining when and where to establish or break contact is critical, yet the number of possible contact sequences grows exponentially with the number of contact modes and the planning horizon; (b) \textit{Nonlinear dynamics and nonconvex geometric constraints}: the planned trajectory must satisfy the system's nonlinear dynamics and geometric constraints such as avoiding self and obstacle collisions. Together, these challenges exacerbate the problem's nonconvexity and computational complexity.

\textbf{Problem statement.} Let $N$ represent the planning horizon with $\enum{N} := \{ 1, 2, \dots, N \}$. Define the state trajectory $\{ x_k \}_{k=0}^N \subset \mathbb{R}^{n_x}$ and the control input trajectory $\{ u_k \}_{k=0}^{N-1} \subset \mathbb{R}^{n_u}$. For contact-rich planning problems, we introduce ``\textit{contact variables}'' $\{ \lambda_k \}_{k=0}^{N-1} \subset \mathbb{R}^{n_\lambda}$, which can be interpreted either as a set of binary contact modes or as continuous contact forces (see examples in \S\ref{sec:exp}). With these definitions, we focus on the following general contact-rich planning problem
\begin{subequations} \label{eq:intro:contact-rich}
    \begin{eqnarray}
     \hspace{-5mm} \min_{\substack{ \left\{ x_k \right\}_{k=0}^N, \left\{ u_k \right\}_{k=0}^{N-1} \\ \left\{ \lam{k} \right\}_{k=0}^{N-1} } } & \displaystyle \ell_N(x_N) + \sum_{k=0}^{N-1} \ell_k(x_k, u_k, \lam{k}) \\
     \subject & x_0 = x_\init \\
        & F_k(x_{k-1}, u_{k-1}, \lam{k-1}, x_{k}) = 0, \ k \in \enum{N}  \\
        & (u_{k-1}, \lam{k-1}, x_k) \in \calC_k, \ k \in \enum{N} 
    \end{eqnarray}
\end{subequations}
where $\ell_k, k = 0, \dots , N$ represents instantaneous loss and terminal loss functions. $F_k$ represents the discretized system dynamics obtained from differential algebraic equations and multiple shooting, possibly involving explicit or implicit contact mode switching. $\calC_k$ imposes various types of constraints on $u_{k-1}, \lam{k-1}, x_k$, including (a) control limits; (b) geometric constraints such as collision avoidance; (c) complementarity constraints related to contact. 
A well-known special case of~\eqref{eq:intro:contact-rich} occurs when the dynamics are linear when fixing $\lam{k}$'s. In such case,~\eqref{eq:intro:contact-rich} can be modeled either as mixed-integer linear/quadratic programming~\cite{ding2020iros-motionplanning-multilegged-mixedinteger,marcucci2020arxiv-warmstart-mixedinteger-mpc} or as linear complementarity problems~\cite{aydinoglu2021tro-stabilization-complementary, yunt2006-opttraj-planning-structure-variant}.

In this paper, we do not assume linearity or convexity but assume (a) $\ell_k$ and $F_k$ are polynomial functions; (b) $\calC_k$ is basic semi-algebraic (\ie described by polynomial constraints). Thus,~\eqref{eq:intro:contact-rich} becomes a \textit{polynomial optimization problem} (POP). \ksc{Formulating robotics problems as polynomial optimization is now well established in the literature \cite{lee2008thesis-computationalgeometricmechanics,yang2022pami-outlierrobust-geometricperception,kang2024wafr-strom,teng2023arxiv-geometricmotionplanning-liegroup}, because 3D rotations and rigid body dynamics expressed in maximal coordinates \cite{brudigam2024variational} admit natural polynomial representations.}

\textbf{Previous methods.}
We briefly review five different methods for solving the contact-rich planning~\eqref{eq:intro:contact-rich}.
(a) \textit{Hybrid MPC}: These methods alternate between contact sequence generation using discrete search~\cite{chen2021iros-traopt-tree-search-multi-contact,wu2020icra-r3t-nonlinear-hybrid,cheng2022icra-contact-mode-quasidynamic,mastalli2020icra-crocoddyl} and continuous-state planning with a fixed sequence.
(b) \textit{Mixed-integer programming}: Contact modes are modeled as binary variables, leading to mixed-integer convex programming (with linear dynamics)~\cite{ding2020iros-motionplanning-multilegged-mixedinteger,marcucci2020arxiv-warmstart-mixedinteger-mpc} or mixed-integer nonconvex programming (with nonlinear dynamics)~\cite{koolen2020arxiv-balance-control-humanoid-nonlinear-centroidal}. These methods scale poorly with the planning horizon, as the worst-case computational complexity grows exponentially with the number of binary variables.
(c) \textit{Dynamics smoothing}: This approach approximates nonsmooth complementarity constraints with smooth surrogate functions, simplifying the problem into a smooth nonlinear programming formulation suitable for local solvers~\cite{chatzinikolaidis2021ral-traopt-contact-rich-implicit-ddp,tassa2014icra-control-limitted-ddp,mordatch2012tog-discovery-complex-behaviors-contact-invariant,tassa2012iros-synthesis-stabilization-online-traopt}. Convex smoothing methods~\cite{pang2023tro-global-planning-contact-rich-quasi-dynamic-contact-models} also exist, at the cost of locally linearizing the dynamics. 
(d) \textit{Contact-implicit methods}: These mainstream methods encode contact modes implicitly through contact forces and complementarity constraints. Numerous local solvers are based on this framework~\cite{aydinoglu2023icra-realtime-multicontact-mpc-admm,yunt2007isdc-combined-continuation-penalty,posa2014ijrr-traopt-directmethod-contact,manchester2020isrr-variational-contact-implicit,yang2024rss-dynamic-on-plam-control-sliding,le2024tro-fast-contact-implicit-mpc}. 
% \hy{Did you cite Zac's bilevel work, which can also be viewed as smoothing?}
However, it is well known that contact-implicit planning problems fail the common constraint qualifications that are crucial for the convergence of numerical solvers.
(e) \textit{Graph of convex sets (GCS)}: As a recently proposed \ksc{powerful} planning framework that explicitly models both discrete and continuous actions, GCS has been extended to contact-rich tasks~\cite{graesdal2024arxiv-tightconvexrelax-contactrich,yang2024arxiv-sdp-linear-piecewise-affine-optimal-control,morozov2024arxiv-multi-query-spp-gcs}. These methods can be viewed as an extension of mixed-integer nonconvex optimization with two-level convex relaxations where level one is a semidefinite relaxation and level two involves inequality multiplication. \ksc{However, in contact-rich motion planning, achieving both tight relaxations and fast solve times remains challenging, and the current GCS framework has yet to fully integrate nonlinear dynamics with geometric constraints.}

% \emph{Can we solve the general contact-rich planning problem~\eqref{eq:intro:contact-rich} to (near) global optimality efficiently?}

\begin{quote}
    \textit{Is it possible to solve the contact-rich planning problem~\eqref{eq:intro:contact-rich} to (near) global optimality efficiently?}
\end{quote}

\textbf{``Sparse'' Moment-SOS hierarchy}. 
Modeling contact-implicit planning as polynomial optimization (POP) in~\eqref{eq:intro:contact-rich} brings both opportunities and challenges. 
\begin{itemize}
    \item \textbf{Opportunities.} Lasserre's hierarchy of moment and sums-of-squares (SOS) relaxations~\cite{lasserre2001siopt-global} provides a principled and powerful machinery for global optimization of POPs through convex relaxations. Particularly, the Moment-SOS hierarchy generates a series of convex \textit{semidefinite programs} (SDPs) with growing sizes whose optimal values provide nondecreasing \emph{lower bounds} that asymptotically converge to the global minimum of~\eqref{eq:intro:contact-rich}. Combined with a feasible (or locally optimal) solution of~\eqref{eq:intro:contact-rich} that provides an \emph{upper bound} to the global minimum, one can compute increasingly tight (small) \emph{(sub)optimality certificates} by measuring the relative gap between the lower bound and the upper bound. To enhance scalability of the hierarchy, ``\emph{sparse}'' Moment-SOS hierarchy has been proposed to exploit sparsity in the POPs, including correlative sparsity~\cite{lasserre2006msc-correlativesparse,huang2024arxiv-sparsehomogenization} and term sparsity~\cite{wang2021siam-tssos, magron23book-sparse} (see more details in \S\ref{sec:general-sparsity-ksc}). Notably, the Julia package \tssos~\cite{magron2021arxiv-julia-tssos} supports automatic sparsity exploitation as long as the user provides a POP formulation. The recent work~\cite{ teng2024ijrr-convex-geometric-motion-planning} in robotics applied \tssos to several motion planning problems and demonstrated that sparse relaxations can deliver small suboptimality gaps. Furthermore, \cite{kang2024wafr-strom} has shown that all trajectory optimization problems exhibit a generic \emph{chain-like} correlative sparsity pattern and designed a GPU-based ADMM SDP solver that achieves significant speedup than off-the-shelf SDP solvers. \emph{Can we directly apply sparse Moment-SOS relaxations to the contact-implicit planning problem~\eqref{eq:intro:contact-rich}?}
    
    \item \textbf{Challenges.} The answer is unfortunately NO, due to three challenges. First, multiple contact modes will make the chain-like correlative sparsity pattern introduced in~\cite{kang2024wafr-strom} too large to be solved efficiently. Second, \tssos allows exploiting more flexible sparsity patterns but its automatic sparsity exploitation operates like a black box---it does not visualize the sparsity patterns being exploited and it is unclear whether robotics-specific domain knowledge can lead to customized sparsity. Third, as reported in~\cite{teng2024ijrr-convex-geometric-motion-planning}, the suboptimality gaps when using \tssos for many smooth planning problems are already large (above $20\%$), not to mention the extra nonsmoothness and combinatorial complexity brought by contact-implicit planning. As shown in~\cite{teng2023arxiv-geometricmotionplanning-liegroup}, \tssos can fail to extract feasible solutions for certain difficult instances of~\eqref{eq:intro:contact-rich}.

\end{itemize}

% have effectively addressed these challenges.

% for polynomial optimization problems (POPs), providing a series of \textit{semidefinite programs} (SDPs) that progressively tighten this gap.  

% \textit{Convex relaxation} offers a robust and efficient approach to solving general nonconvex problems by computing a \textit{suboptimality gap} through a \textit{lower bound} from the convex relaxation and an \textit{upper bound} from a feasible solution of the nonconvex problem. Lasserre's Moment and Sum-of-Squares (SOS) Hierarchy~\cite{lasserre2001siopt-global} is a primary method for polynomial optimization problems (POPs), providing a series of \textit{semidefinite programs} (SDPs) that progressively tighten this gap. 

% However, scalability becomes problematic for planning problems, as polynomial variable counts can surge into the thousands. Advances in sparse Moment-SOS Hierarchy, including correlative sparsity~\cite{lasserre2006msc-correlativesparse} and term sparsity~\cite{wang2021siam-tssos, magron23book-sparse}, have effectively addressed these challenges. 

% \textbf{Sparse Moment-SOS Hierarchy: Challenges.} 

\textbf{Contributions.} 
In this paper, we tackle the aforementioned challenges and show that it is indeed possible to solve many instances of the contact-implicit planning problem~\eqref{eq:intro:contact-rich} to near global optimality. The key strategy is to build ``sparsity-rich'' semidefinite relaxations from the ground up, for robotics. 

We summarize our contributions as follows. 
% globally solve the contact-rich planning problem described in Equation~\eqref{eq:intro:contact-rich} using sparsity-rich semidefinite relaxation. 
\begin{enumerate}[label=(\Roman*)]
    \item \textbf{White-box sparsity exploitation.} 
    We provide a tutorial-style review of the fundamental mathematical concepts underpinning correlative and term sparsity for POPs, and further ground our discussion in a concrete contact-implicit planning problem. 
    We build a new C++ Sparse Polynomial Optimization Toolbox (\spot), interfacing both Matlab and Python, that (a) is faster than \tssos, (b) offers richer relaxation options, and (c) visualizes the automatically discovered sparsity patterns (see Fig.~\ref{fig:demos}).
    % We comprehensively review the theorems and algorithms for correlative and term sparsity clique generation. As a result, a new C++ sparse POP conversion package, \spot, is developed, offering faster conversion speeds and richer conversion options compared to TSSOS~\cite{magron2021arxiv-julia-tssos}.
    \item \textbf{Robotics-specific sparsity.} Beyond automatic exploitation of the generic correlative and term sparsity, we show that it is possible and crucial to exploit robotics-specific sparsity patterns. Particularly, we investigate sparsities derived from robot kinematic chains and separable contact modes, and demonstrate that robotics-specific sparsity patterns achieve both tighter lower bounds and reduced computation times compared to automatically generated ones in large-scale problems such as Planar Hand. 
    \item \textbf{Robust minimizer extraction.} 
    An important but often overlooked problem in SDP relaxations is how to extract good solutions to the nonconvex optimization from optimal SDP solutions, especially when the relaxation is not tight (\ie the suboptimality gap is large).  
    Inspired by the recent advances in Gelfand-Naimark-Segal (GNS) construction~\cite{klep2018siopt-minimizer-extraction-robust}, we develop a new minimizer extraction routine for sparse Moment-SOS relaxations that demonstrates superior robustness over naive extraction methods previously implemented in~\cite{kang2024wafr-strom,magron2021arxiv-julia-tssos}.
    \item \textbf{Extensive case studies.} We test our sparse semidefinite relaxations on five contact-rich planning problems: Push Bot, Push Box, Push T, Push Box with Obstacles, and Planar Hand. 
    Of independent interests, some of our polynomial modeling techniques also appear to be new in the planning literature. 
    Thanks to rich sparsity, the generated small-to-medium scale SDP relaxations can be solved in \textit{seconds} while achieving decent tightness and certified global optimality.
    Furthermore, we showcase robust push-T performance of our SDP relaxations using a real-world robotic manipulator. In fact, with global optimization, model predictive control is so robust that it succeeds the task even under severe environment disturbances that effectively make the ``model wrong'' (see Fig.~\ref{fig:demos}). 
\end{enumerate}

\textbf{Paper organization.} We present correlative and term sparsity in \S\ref{sec:general-sparsity-ksc}, robotics-specific sparsity in \S\ref{sec:robotics-specific}. We give numerical and real-world experiments in \S\ref{sec:exp}, and conclude in \S\ref{sec:conclusion}.

\textbf{Notations.} Let $\mvx = (x_1, \dots, x_n)$ be a tuple of variables and $\R[\mvx] = \R[x_1,\dots,x_n]$ be the set of polynomials in $\mvx$ with real coefficients. A monomial is defined as $\mvx^\mva = x_1^{\alpha_ 1}x_2^{\alpha_2} \cdots x_n^{\alpha_n}$. A polynomial in $\mvx$ can be written as $f(\mvx) = \sum_{\mva \in \bbN^n} f_\mva \mvx^\mva$ with coefficients $f_\mva \in \R$. We denote the set of all polynomials with degree less than or equal to $d$ as $\R_{d}[\mvx]$. 
The support of $f$ is defined by $\text{supp}(f) = \left\{ \mva\in \bbN^n\mid f_\mva \ne 0 \right\}$, i.e., the set of exponents with nonzero coefficients. The set of all variables contained in $f$ is defined by $\text{var}(f)$.
Let $\mvx^{\bbN_d^n}$ be the \emph{standard monomial basis}, abbreviated as ${[\mvx]}_d$. Given an index set $I\subseteq [n]$, let $\mvx(I)= (x_i, i\in I)$ and ${[\mvx(I)]}_d$ denote the standard monomial basis of the subspace spanned by the variables $x_i, i\in I$.
An undirected graph $G(V, E)$ consists of a vertex set $V={v_1, v_2, \dots, v_n}$ and an edge set $E \subseteq \{(v_i, v_j) \mid v_i, v_j\in V, v_i\ne v_j \}$. 
Let $\mbS^n$ denote the space of $n \times n$ symmetric matrices, and $\mbS^n_+$ denote the cone of $n \times n$ symmetric positive semidefinite (PSD) matrices.

% \input{sections/related-works.tex}
%!TEX root = ../main.tex

\section{Correlative and Term Sparsity}
\label{sec:general-sparsity-ksc}

In this section, we review a systematic mechanism to relax a  general (nonconvex) polynomial optimization problem (POP) as a (convex) semidefinite program (SDP) while exploiting two levels of sparsity: (a) variable level---correlative sparsity (CS); and (b) term level---term sparsity (TS). 
Formally, we consider the following POP
\begin{align}
    \label{eq:gs:general-pop}
    \rho^\star = \min_{\mvx\in \R^n}\left\{
        f(\mvx)\mymid  \substack{ \displaystyle g_1(\mvx)\ge 0, \dots, g_{m_{\ineq}}(\mvx)\ge 0,\\
        \displaystyle h_1(\mvx) = 0, \dots, h_{m_{\eq}}(\mvx) = 0}
    \right\} 
\end{align}
where the objective function $f$ and the constraints $g_i, h_i$ are all real polynomials. To ground our discussion in a concrete robotics example, let us consider the following simple contact-rich motion planning problem.

\begin{figure}[htbp]
    \centering
    \begin{minipage}{\columnwidth}
        \centering
        \begin{tabular}{c}
            \begin{minipage}{0.5\columnwidth}
                \centering
                \includegraphics[width=\columnwidth]{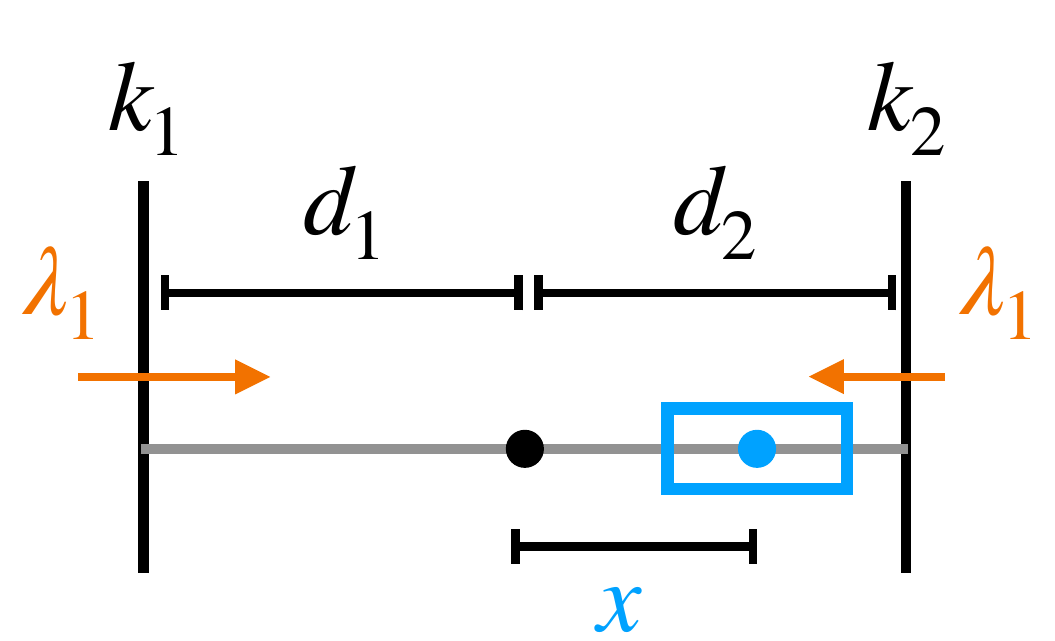}
            \end{minipage}
        \end{tabular}
    \end{minipage}

    \caption{Double integrator with soft wall.
    \label{fig:sp:double-integrator}}
    \vspace{-6mm}
\end{figure}

\begin{example}[Double Integrator with Soft Wall]
	\label{exa:gs:di-soft-wall}
    As shown in Fig.~\ref{fig:sp:double-integrator}, consider a point mass $m$ driven by a control force $u$ that can bounce between two soft walls with spring coefficients $k_1$ and $k_2$. Denote the system state as $(x, v)$ ($x$: position, $v$: velocity), the control input as $u$, and the two wall's forces as $(\lam{1}, \lam{2})$, we consider the following trajectory optimization (optimal control) problem
    \begin{subequations}
        \begin{align}
            \min \quad & \sum_{k = 0} ^ {N - 1} u_k^2 + \x[k+1]^2 + v_{k+1}^2 \label{eq:di:obj}\\
            \text{s.t.} \quad & \x[k+1] - \x[k] = \dt \cdot v_k \label{eq:di:dynamics:1}\\
            & v_{k+1} - v_k = \frac{\dt}{m} \cdot (u_k + \lam{1}[k] - \lam{2}[k]) \label{eq:di:dynamics:2} \\
            & u_{\max}^2 - u_k^2 \ge 0 \label{eq:di:control}\\
            & 0 \le \lam{1}[k] \perp \frac{\lam{1}[k] }{k_1} + d_1 + \x[k] \ge 0 \label{eq:di:force:1}\\
            & 0 \le \lam{2}[k] \perp \frac{\lam{2}[k] }{k_2} + d_2 - \x[k] \ge 0 \label{eq:di:force:2} \\
            & x_0 = x_{\init} \quad \text{and} \quad v_0 = v_{\init} \label{eq:di:init}
        \end{align} 
    \end{subequations}
where $\dt$ represents the time discretization, \eqref{eq:di:obj} formulates a quadratic regulation loss function around $x=0,v=0$, \eqref{eq:di:dynamics:1}-\eqref{eq:di:dynamics:2} represent system dynamics, \eqref{eq:di:control} enforces control saturation, \eqref{eq:di:force:1}-\eqref{eq:di:force:2} specify the soft complementarity constraints between position and contact forces, and \eqref{eq:di:init} provides the initial condition. $\lam{1}[k]$ denotes the contact force at step $k$.
\end{example}

\textbf{Outline.} We begin by introducing the fundamentals of chordal graphs, a key mathematical tool for analyzing sparsity patterns (\S\ref{sec:gs:chordal-graph}). Next, we review correlative sparsity (CS) (\S\ref{sec:gs:cs}) and term sparsity (TS) (\S\ref{sec:gs:ts}) separately. Finally, we introduce our high-performance C++ sparse polynomial optimization toolbox, \spot (\S\ref{sec:gs:spot}). 

% To complement these mathematical concepts, we present a toy example — the double integrator with soft walls — to illustrate the two-level sparsity patterns inherent in contact-rich planning problems.

%!TEX root = ../../main.tex
\vspace{-2mm}

\subsection{Chordal Graph}
\label{sec:gs:chordal-graph}

\begin{definition}[Chordal graph]
    A graph $G = (V, E)$ is \emph{chordal} if every cycle of four or more vertices has a chord---an edge connecting two non-adjacent vertices in the cycle. 
    
    % Formally, for any induced cycle $C \subseteq V$ with $|C| \geq 4$, there exists an edge $(v_i, v_j) \in E$ such that $v_i, v_j \in C$ and $(v_i, v_j)$ is not part of the cycle $C$.
\end{definition}

\begin{figure}[htbp]
    \centering
    \includegraphics[width=0.9\columnwidth]{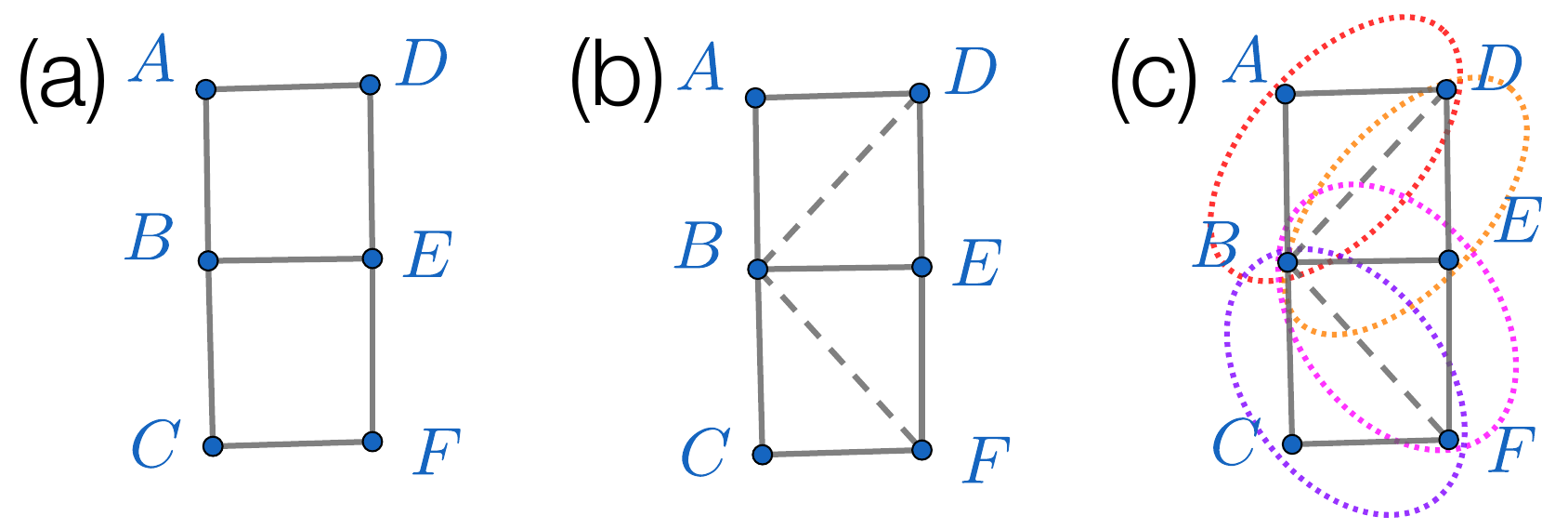}
    \caption{An example of chordal extension and maximal cliques.
    \label{fig:gs:chordal-extension-ksc}}
    \vspace{-8mm}
\end{figure}

For a quick example, the graph in Fig.~\ref{fig:gs:chordal-extension-ksc}(a) is not chordal but the graph in Fig.~\ref{fig:gs:chordal-extension-ksc}(b) is. Apparently, one can change a non-chordal graph to chordal by adding edges. This is the notion of a chordal extension.

\textbf{Chordal extension.} A graph $G'(V', E')$ is called a \emph{chordal extension} of graph $G(V, E)$ if (a) $G'$ is chordal, and (b) $V' = V$ and $E\subseteq E'$. The ideal objective of chordal extension is to add the \emph{minimum} number of edges to $G$ to make $G'$ chordal. However, it is known that finding such a minimal chordal extension is NP-complete~\cite{Yannakakis1981siam-minimum-fill-in}.
% \footnote{A minimal chordal extension refers to adding the minimum number of edges to the graph $G$ such that the resulting graph $G'$ is chordal. This problem is also known as the Minimum Fill-In problem.} 
A common heuristic for approximating the minimal chordal extension is the \emph{minimal degree} (MD) chordal extension~\cite{Rose1976siam-vertex-elimination}. An alternative heuristic is to select vertices based on the number of additional edges (MF) required to maintain chordality~\cite{Yannakakis1981siam-minimum-fill-in}. We summarize the MD chordal extension method in Algorithm~\ref{alg:gs:md} and MF chordal extension in Algorithm~\ref{alg:gs:mf} in Appendix~\ref{app:sec:mdmf}. 
 In our \spot package, we implement both algorithms.

\textbf{Maximal cliques.} Once the chordal extension is constructed, the next step is to identify the \emph{maximal cliques}. 

\begin{definition}[Clique]
    A \emph{clique} in a graph $G(V, E)$ is a subset of vertices $C \subseteq V$ where every pair of vertices is connected by an edge. 
    A clique is \emph{maximal} if it is not properly contained within any other clique in $G$.
\end{definition}

The nice consequence of the chordal extension is that the maximal cliques of a chordal graph can be enumerated efficiently in linear time in terms of the number of nodes and edges~\cite{Delbert1965pjm-matrix-and-graph, Hans2010iac-treewidth-computations, golumbic2004algorithmic}. 
Identifying all maximal cliques in a chordal graph plays a fundamental role in exploiting sparsity patterns.

%!TEX root = ../../main.tex

\begin{example}[Chordal Extension and Maximal Cliques]
    Consider the graph in Fig.~\ref{fig:gs:chordal-extension-ksc}(a) that is non-chordal. The MD chordal extension Algorithm~\ref{alg:gs:md} adds two edges $(B,D)$, $(B,F)$, leading to the chordal graph in Fig.~\ref{fig:gs:chordal-extension-ksc}(b). 
    The maximal cliques of the resulting chordal graph are $\cbrace{A,B,D}$, $\cbrace{B,D,E}$, $\cbrace{B,E,F}$, and $\cbrace{B,C,F}$, as shown in Fig.~\ref{fig:gs:chordal-extension-ksc}(c).
\end{example}

With the notion of a chordal graph and maximal cliques, it is natural to use such a graph-theoretic tool to exploit sparsity.

\subsection{Correlative Sparsity}
\label{sec:gs:cs}

As mentioned before, correlative sparsity (CS) seeks to exploit sparsity in the ``variable'' level. Roughly speaking, the intuition is that we can construct a graph that represents the connectivity in the POP~\eqref{eq:gs:general-pop}, perform a chordal extension to that graph, and find its maximal cliques to group the (potentially very large number of) POP variables into many groups where each group only contains a few variables~\cite{Waki2006siam-sos-semidefinite-relaxation}.

% Correlative sparsity (CS) pertains to the variables in a POP and is leveraged by partitioning and regrouping these variables into cliques based on their interconnections. There are two steps: (a) CS graph construction; and (b) chordal extension and regrouping. 

\textbf{CS graph construction.}  The graph $G^{\csp}(V, E)$ is the correlative sparsity pattern (CSP) graph of a POP with variables $\mvx \in \R^n$ if $V = [n]$ and $(i, j) \in E$ if at least one of the following three conditions holds:
\begin{enumerate}
    \item $\exists \mva \in \text{supp}(f)$, s.t. $\alpha_i, \alpha_j > 0$,
    \item $\exists k \in [m_{\ineq}]$, s.t. $x_i, x_j \in \text{var}(g_k)$,
    \item $\exists k \in [m_{\eq}]$, s.t. $x_i, x_j \in \text{var}(h_k)$.
\end{enumerate}

In words, the nodes of the CSP graph represent variables of the POP, and a pair of nodes are connected if and only if they simultaneously appear in the objective or the constraints.

\textbf{Chordal extension and grouping.} With the MD chordal extension Algorithm~\ref{alg:gs:md} (or the MF chordal extension Algorithm~\ref{alg:gs:mf}), we compute the chordal extension of $G^{\csp}$--- denoted $(G^{\csp})'$---and its maximal cliques $\cbrace{I_l}_{l=1}^{p}$, where each clique $I_l$ contains a set of nodes (variables). We then partition the constraint polynomials $g_1, \dots, g_{m_{\ineq}}$ and $h_1, \dots, h_{m_{\eq}}$ into groups $\left\{ g_j\mymid j\in \calG_l \right\}$ and $\left\{ h_j\mymid j\in \calH_l \right\}$, where $\mathcal{G}_l$ and $\mathcal{H}_l$, $l \in \enum{p}$ index the inequality and equality constraints involving the variables in $I_l$, respectively. Formally, this is 
\begin{enumerate}
    \item $\forall j \in \calG_l, \text{var}(g_j)\subseteq I_l$, 
    \item $\forall j \in \calH_l, \text{var}(h_j)\subseteq I_l$.
\end{enumerate}

%!TEX root = ../../main.tex

\begin{figure}[htbp]
    \centering
    \begin{minipage}{\columnwidth}
        \centering
        \begin{tabular}{c}
            \begin{minipage}{\columnwidth}
                \centering
                \includegraphics[width=\columnwidth]{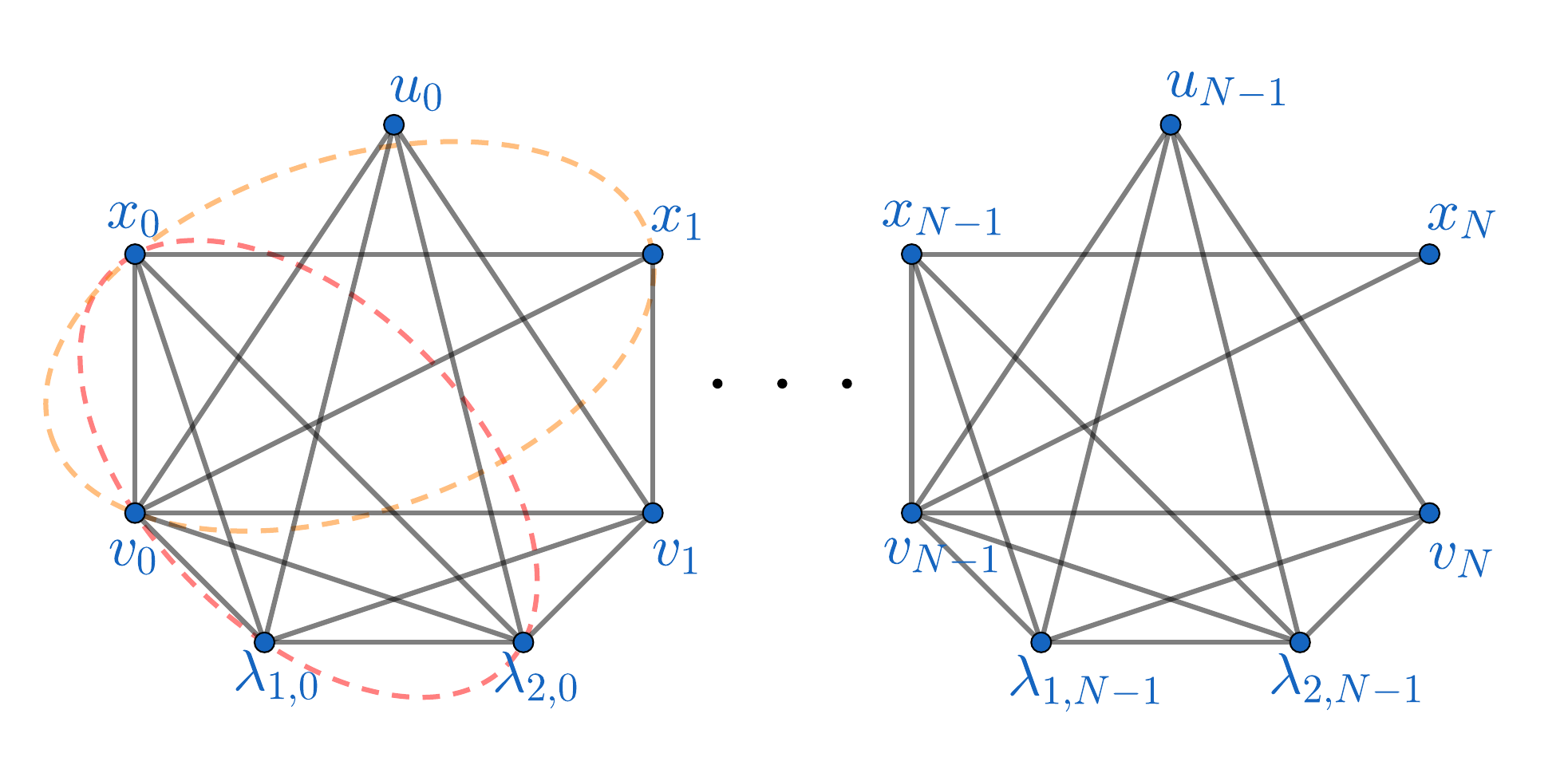}
            \end{minipage}
        \end{tabular}
    \end{minipage}

    \caption{CSP graph of the toy Example~\ref{exa:gs:di-soft-wall}. Red circle: maximal clique $\cbrace{x_0, v_0, \lambda_{1,0}, \lambda_{2,0}}$. Orange circle: maximal clique $\cbrace{x_0, x_1, v_0}$. Only variables in the first and last planning steps are shown for simplicity.
    \label{fig:sp:csp-graph}}

    \vspace{-3mm}
\end{figure}

We make this concrete by recalling our robotics Example~\ref{exa:gs:di-soft-wall}.

\begin{example}[CSP Graph of Example~\ref{exa:gs:di-soft-wall}]
	Fig.~\ref{fig:sp:csp-graph} shows the CSP graph of Example~\ref{exa:gs:di-soft-wall}. It is already a chordal graph without chordal extension. Two maximal cliques are highlighted: $\cbrace{x_0, v_0, \lambda_{1,0}, \lambda_{2,0}}$ and $\cbrace{x_0, x_1, v_0}$. There is an edge between $x_0$ and $v_0$ since $x_1 - x_0 = \dt \cdot v_0$ shows up in~\eqref{eq:di:dynamics:1}.
\end{example}

After grouping the variables into cliques $\{I_l\}_{l=1}^p$ and the polynomial constraints into subsets $\calH_l,\calG_l$, we can design a hierarchy of ``sparse'' moment and sums-of-squares (SOS) relaxations to globally optimize the POP~\eqref{eq:gs:general-pop}. 

Before we present the sparse Moment-SOS hierarchy, it is useful to understand the ``dense'' Moment-SOS hierarchy.

\textbf{Dense Moment-SOS relaxations.} Recall the POP~\eqref{eq:gs:general-pop}. Let $\mvy = (y_\mva)_{\mva}$ be a sequence of real numbers indexed by the standard monomial basis of $\R[\mvx]$. Define the Riesz linear functional $L_{\mvy} \colon \R[\mvx]\to \R$ as:
\begin{align}
    f = \sum_{\mva}f_{\mva}\mvx^\mva \mapsto \sum_{\mva}f_{\mva}y_{\mva}, \quad \forall f(\mvx) \in \R[\mvx]
\end{align}
In words, the Riesz linear functional $L_\mvy$ transforms a real polynomial $f$ to a real number that is the inner product between $\mvy$ and the vector of coefficients of $f$. The notation of $L_{\mvy}$ can be naturally extended to polynomial vectors and matrices, as illustrated in the following example.
\begin{example}[Riesz Linear Functional]
    Let $n = 3$, $I = \left\{ 1,3 \right\}$. Then $\sqbk{\mvx}_1 = \left[1;x_1;x_2;x_3\right]$ and $\sqbk{\mvx(I)}_2 = \left[ 1;x_1;x_3;x_1^2;x_1x_3;x_3^2 \right]$. Applying $L_{\mvy}$, we have
    \begin{align}
        & L_{\mvy}((x_1 + x_3)\cdot \sqbk{\mvx(I)}_2) = 
	 \sqbk{
		\begin{array}{c}
			y_{1,0,0} + y_{0,0,1}\\
			y_{2,0,0} + y_{1,0,1}\\
			y_{1,0,1} + y_{0,0,2}\\
			y_{3,0,0} + y_{2,0,1}\\
			y_{2,0,1} + y_{1,0,2}\\
			y_{1,0,2} + y_{0,0,3}
		\end{array}
     },
    \end{align}
    \begin{align}
	 & L_{\mvy}(\sqbk{\mvx}_1\sqbk{\mvx}_1^T) = 
		\sqbk{
			\begin{array}{cccc}
				y_{0,0,0} & y_{1,0,0} & y_{0,1,0} & y_{0,0,1}\\
				y_{1,0,0} & y_{2,0,0} & y_{1,1,0} & y_{1,0,1}\\
				y_{0,1,0} & y_{1,1,0} & y_{0,2,0} & y_{0,1,1}\\
				y_{0,0,1} & y_{1,0,1} & y_{0,1,1} & y_{0,0,2}
			\end{array}
		},
    \end{align}
    where the number $y_{1,0,0}$ is applying $L_\mvy$ to the monomial $x_1^1 \cdot x_2^0 \cdot x_3^0 = x_1$.
\end{example} 

With the Riesz linear functional, we can state the dense Moment-SOS hierarchy. Essentially, through the Riesz linear functional $L_\mvy$, the Moment-SOS hierarchy relaxes the original POP~\eqref{eq:gs:general-pop} as a convex optimization problem whose variable becomes the sequence $\mvy$. The reason why this is called a ``hierarchy'' is because one can make the sequence arbitrarily long, depending on how many monomials are included. 

\begin{proposition}[Dense Moment-SOS Hierarchy]\label{prop:dense}
    Consider the POP~\eqref{eq:gs:general-pop}. Let $d_j^g = \ceil{\text{deg}(g_j)/2}$, $\forall j \in \enum{m_{\ineq}}$ and $d_j^h = \text{deg}(h_j)$, $\forall j \in \enum{m_{\eq}}$. Define
    \bea\label{eq:min-relaxation-order} 
    \hspace{-2mm}d_{\min}\!\!=\!\!\max\cbrace{ \ceil{\text{deg}(f)/2}, \cbrace{d_j^g}_{j\in \enum{m_{\ineq}}}, \cbrace{\ceil{d_j^h / 2}}_{j \in \enum{m_\eq}} }.\!\!\!\!
    \eea
    Given a positive integer $d \geq d_\min$, 
    the $d$-th order dense Moment-SOS hierarchy reads:
    \begin{subequations}\label{eq:gs:ds-moment}
        \begin{eqnarray}
            \min_\mvy & L_\mvy(f) \\
            \text{s.t.} & L_\mvy\left( \enum{\mvx}_d \enum{\mvx}_d\tran \right) \succeq 0\\
            & L_\mvy\left( g_j \cdot\enum{\mvx}_{d-d_j^g} \enum{\mvx}_{d-d_j^g}\tran \right) \succeq 0, \forall j\in \calG_l, l\in [p]\\
            & L_\mvy\left( h_j \cdot \enum{\mvx}_{2d-d_j^h} \right) = 0, \forall j\in \calH_l, l\in [p]\\
            & \mvy_{\mathbf{0}} = 1 
        \end{eqnarray}
    \end{subequations}
    The optimal value of the convex optimization~\eqref{eq:gs:ds-moment} converges to the optimal value of the nonconvex POP~\eqref{eq:gs:general-pop} $\rho^\star$ as $d \rightarrow \infty$.
    % \hy{@Shucheng, the $d_\min$ definition seems wrong, and please write the dense version similar as the sparse version below.}
\end{proposition}
In~\eqref{eq:gs:ds-moment}, the matrix $L_\mvy([\mvx]_d [\mvx]_d\tran)$ is usually called a \emph{moment matrix} and it is enforced by a positive semidefinite (PSD) constraint.
One can see that the dense Moment-SOS hierarchy can become expensive very quickly as the relaxation order $d$ increases. This is because the ``dense'' moment matrix at order $d$ has size $(\substack{n+d\\d})$ which quickly makes the PSD constraint too large to be handled by off-the-shelf SDP solvers (recall that the length of the monomial basis indexing the moment matrix---$[\mvx]_d$---is $(\substack{n+d\\d})$).

\textbf{Moment-SOS relaxations with CS.} On the other hand, with correlative sparsity and when the variables are divided into cliques $\{I_l\}_{l=1}^p$ where the size of clique $I_l$ is $n_l$, then instead of generating a single moment matrix with size $\nchoosek{n+d}{d}$, the sparse Moment-SOS hierarchy will generate $p$ moment matrices where the size of each moment matrix is $\nchoosek{n_l+d}{d}$. A different way to view this is that, correlative sparsity breaks a large PSD constraint into multiple smaller PSD constraints. 

Let us formalize this.

\begin{proposition}[Sparse Moment-SOS Hierarchy]\label{prop:sparse} Consider the POP~\eqref{eq:gs:general-pop} and assume its variables are grouped into cliques $\{ I_l\}_{l=1}^p$ and its constraints are grouped into $\calG_l,\calH_l$ where $\calG_l$ and $\calH_l$ include constraints only involving variables $\mvx[I_l]$. For any fixed integer $d \geq d_\min$ ($d_\min$ defined as in~\eqref{eq:min-relaxation-order}), define the following polynomial matrices and vectors associated with each clique $I_l$ as:
    \begin{subequations}\label{eq:sparse-matrices}
        \begin{align}
            M_d(I_l) & = {\sqbk{\mvx(I_l)}_d} \sqbk{\mvx(I_l)}_d\tran, l\in \sqbk{p}\\
            M_d(g_j, I_l) & = g_j\cdot {\sqbk{\mvx(I_l)}_{d-d_j^g}} \sqbk{\mvx(I_l)}_{d-d_j^g}\tran, j\in \calG_l, l\in \sqbk{p}\\
            H_d(h_j, I_l) & = h_j\cdot \sqbk{\mvx(I_l)}_{2d-d_j^h}, j\in \calH_l, l\in \sqbk{p} \label{eq:gs:H}
        \end{align}
    \end{subequations}
    Let  $g_0 := 1$, then $M_d(g_0, I_l) = M_d(I_l)$. The $d$-th order Moment-SOS hierarchy with correlative sparsity reads:
    \begin{subequations}\label{eq:gs:cs-moment}
        \begin{eqnarray}
            \hspace{-5mm}\rho_d := \min_\mvy & L_\mvy(f) \\
            \text{s.t.} & L_\mvy\left( M_d(g_j, I_l) \right) \succeq 0, \forall j\in \cbrace{0} \cup \calG_l, l\in [p]\\
            & L_\mvy\left( H_d(h_j, I_l) \right) = 0, \forall j\in \calH_l, l\in [p]\\
            & \mvy_{\mathbf{0}} = 1 
        \end{eqnarray}
    \end{subequations}
    Moreover, under appropriate compactness assumptions~\cite{Lasserre2006siam-convergent-sdp-relaxation}, the sequence ${\rho_d} \rightarrow \rho^\star$ as $d \rightarrow \infty$.
\end{proposition}

%!TEX root = ../../main.tex

\begin{figure}[htbp]
    \centering
        \includegraphics[width=0.9\columnwidth]{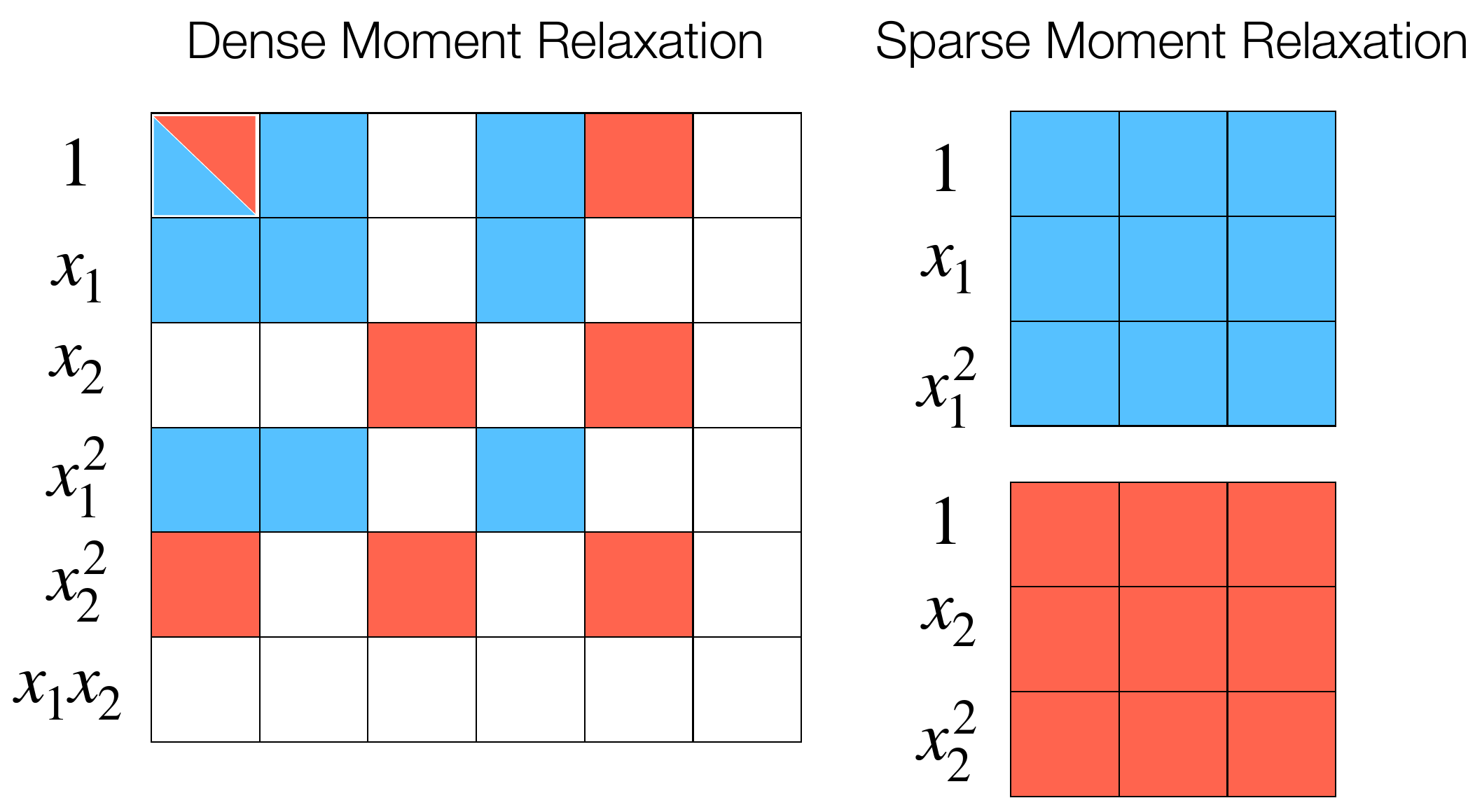}
    \caption{Comparison of the moment matrices in dense and sparse Moment-SOS relaxations.
    \label{fig:sp:compare-dense-sparse}}
    \vspace{-6mm}
\end{figure}

As an illustrative example, suppose the POP has two variables $\mvx=(x_1,x_2)$. Then the moment matrix corresponding to the dense Moment-SOS hierarchy at $d=2$ is shown in Fig.~\ref{fig:sp:compare-dense-sparse} left, where its rows and columns are indexed by the standard monomial basis $[\mvx]_2$. However, suppose from the chordal extension of the CSP graph one can find two cliques $\{x_1\}$ and $\{x_2\}$ (i.e., they are not connected), then the sparse Moment-SOS hierarchy at $d=2$ would generate two moment matrices shown in Fig.~\ref{fig:sp:compare-dense-sparse} right. From the color coding in Fig.~\ref{fig:sp:compare-dense-sparse}, we can see that the two smaller moment matrices are principal submatrices of the big moment matrix. Hence, a $6 \times 6$ PSD constraint is broken into two $3 \times 3$ PSD constraints.

It is worth mentioning that (a) while both the dense and the sparse Moment-SOS hierarchy converge to $\rho^\star$, they may, and in general, converge at different speeds; (b) associated with both~\eqref{eq:gs:ds-moment} and~\eqref{eq:gs:cs-moment} are their dual sums-of-squares (SOS) convex relaxations (hence the name Moment-SOS hierarchy). Though we do not explicitly state the SOS relaxations (see~\cite{Yang2024book-sdp} and references therein), our \spot package implements them.

%%%%%%%%%%%%%%%%%%%%%%% Term sparsity %%%%%%%%%%%%%%%%%%%%%%%

\subsection{Term Sparsity}
\label{sec:gs:ts}

While correlative sparsity (CS) focuses on relationships between variables, term sparsity (TS) addresses relationships between monomials. Specifically, TS is designed to partition monomial bases into blocks according to the monomial connections in the POP~\cite{wang2021siam-tssos}. Rather than analyzing general TS in isolation, we focus on the integration of CS and TS~\cite{wang2022tms-cs-tssos}.

Similar to CS, exploiting TS is also related to constructing a graph called the term sparsity pattern (TSP) graph. This construction involves three steps: (a) initialization; (b) support extension; and (c) chordal extension.

We remark that the notion of TS and the construction of the TSP graph can be less intuitive than the CSP graph mentioned before, and the mathematical notations can get quite involved. However, it is safe for the reader to quickly glance the TSP construction just to understand its high-level idea, and revisit the math a couple more times later.

\textbf{Initialization.} Let ${I_l}, \ l\in [p]$ be the maximal cliques of ($G^{\csp})'$, with $n_l \coloneqq |I_l|$ the size of each clique. Let $\mathcal{G}_l$ and $\mathcal{H}_l$, $l\in [p]$ be defined in the CS grouping procedure and contain polynomial constraints related to clique $I_l$. The variables $\mvx$ are grouped into subsets $\mvx(I_1), \dots, \mvx(I_l)$. Denote $\calA$ as the set of all monomials appearing in the POP~\eqref{eq:gs:general-pop}, union with all even-degree monomials:
\begin{align}
	\label{eq:gs:ts-A}
	\calA \coloneqq \text{supp}(f)\cup\bigcup_{j = 1}^{m_{\ineq}}\text{supp}(g_j)\cup\bigcup_{j = 1}^{m_{\eq}}\text{supp}(h_j)\cup (2\bbN)^n
\end{align}
where $(2\bbN)^n$ is defined as $\cbrace{2\mva \mymid \mva\in \bbN^n}$.

\textbf{Support extension.} For each $l \in [p]$ and $j \in \{0\} \cup \mathcal{G}_l$, construct constraint $g_j$'s TSP graph $G_{d,l,j}^g$ with:
\begin{enumerate}
    \item Nodes: $V_{d,l,j} := |[\mathbf{x}(I_l)]_{d-d_j^g}|$ (these relate to monomials)
    \item Edges: $E_{d,l,j} = \cbrace{(\beta, \gamma)\mymid \text{supp}( M_d(g_j, I_l)_{\beta,\gamma})\cap \calA \ne 0}$, where 
    \begin{align}
        \label{eq:gs:supp-beta-gamma}
       \hspace{-4mm} M_d(g_j, I_l)_{\beta,\gamma} = g_j \cdot [\mathbf{x}(I_l)]_{d-d_j^g}(\beta) \cdot [\mathbf{x}(I_l)]_{d-d_j^g}(\gamma).
    \end{align}
\end{enumerate}
For each each $l \in [p]$ and $j \in \mathcal{H}_l$, define the binary mask vector $B_{d,l,j}^h$ as:
\begin{align}
	\label{eq:gs:ts-equality-support-extension}
	B_{d,l,j}^h(\beta) = \begin{cases}
		1, & \text{supp}(H_d(h_j, I_l)_{\beta}) \cap \calA \ne \emptyset\\
		0, & \text{otherwise}
	\end{cases}
\end{align}
where 
\begin{align}
    H_d(h_j, I_l)_{\beta} = h_j \cdot [\mvx(I_l)](\beta).
\end{align}

\textbf{Chordal extension.} For each $l \in [p]$ and $j \in \{0\} \cup \mathcal{G}_l$, let $G'_{d,l,j}$ be the chordal extension of $G_{d,l,j}^g$. Define $B_{d,l,j}^g$ as its adjacency matrix, which is naturally a binary mask matrix.

%!TEX root = ../../main.tex

\begin{figure}[htbp]
    \centering
    \includegraphics[width=0.9\columnwidth]{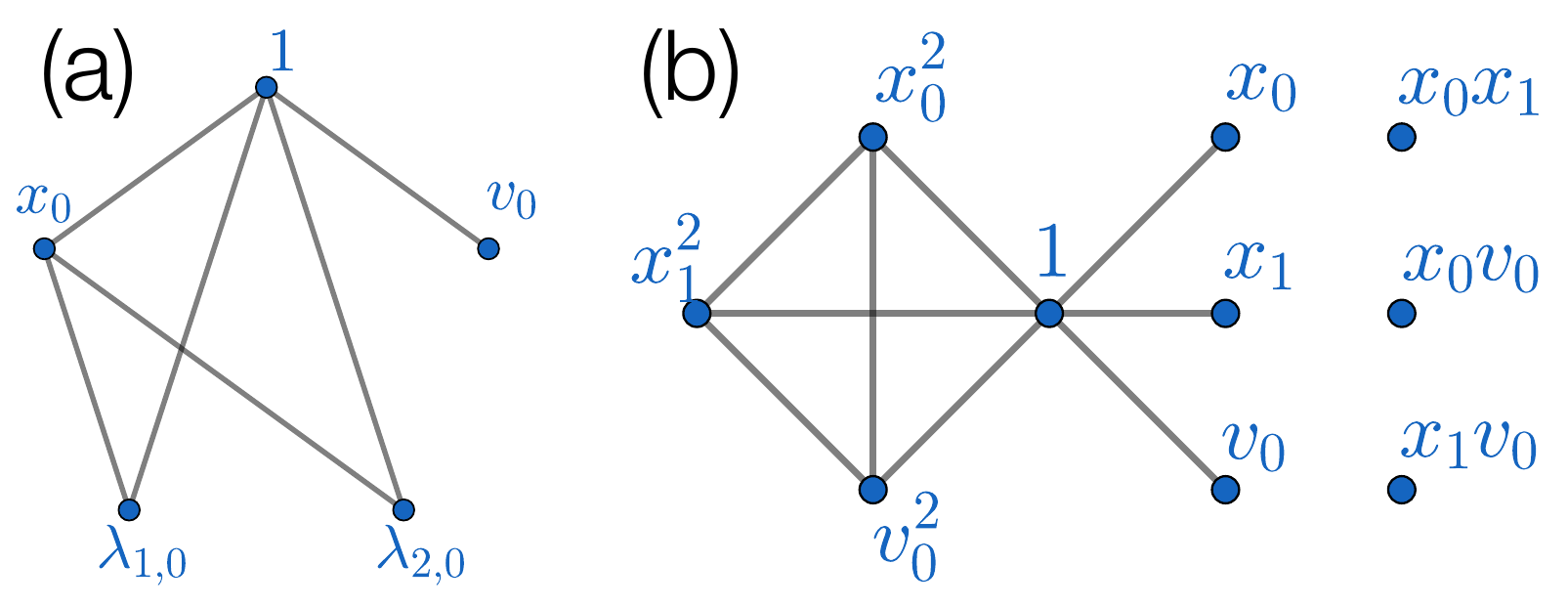}
    \caption{(a) $G_{2,1,1}^g$: support extension for $g_1$ in clique $I_1$; (b) $G_{2,2,0}^g$: support extension for $g_0$ in clique $I_2$. \label{fig:sp:support-extension}}
    \vspace{-3mm}
\end{figure}

We shall illustrate this using our robotics example.

\begin{example}[TSP Graph of Example~\ref{exa:gs:di-soft-wall}]
	Consider the toy example's two cliques: $I_1 := \cbrace{x_0, v_0, \lambda_{1,0}, \lambda_{2,0}}$ and $I_2 := \cbrace{x_0, x_1, v_0}$ (illustrated in Fig.~\ref{fig:sp:csp-graph}). Define $g_1 := \lambda_{1,0} + 1 + x_0$ and $g_0 := 1$. Then, $G_{2,1,1}^g$ is illustrated in Fig.~\ref{fig:sp:support-extension}(a). 
    
    There exists an edge between $x_0$ and $\lam{1,0}$ since $x_0$ and $\lam{1,0}$ appear at the second and fifth positions of $\mvx(I_1)_{d - d_g} := [1, x_0, v_0, \lambda_{1,0}, \lambda_{2,0}]\tran$, respectively. According to~\eqref{eq:gs:supp-beta-gamma}, 
    \begin{align}
        \lambda_{2,0} x_0 \in M_2(g_1, I_1)_{2,5} = (\lambda_{1,0} + 1 + x_0) \cdot x_0 \cdot \lambda_{2,0} 
    \end{align}
    and $\lambda_{2,0} x_0 \in \calA$ since $\lam{2}[0] \cdot (\frac{\lam{2}[0] }{k_2} + d_2 - \x[0]) = 0$. 
    
    Similarly, $G_{2,2,0}^g$ is illustrated in Fig.~\ref{fig:sp:support-extension}(b). There exists an edge between $x_0^2$ and $x_1^2$ since $x_0^2 \cdot x_1^2$ is of even degree. Both $G_{2,1,1}^g$ and $G_{2,2,0}^g$ are already chordal. 
\end{example}

Note that the crucial difference between CSP and TSP is that the nodes of a CSP graph are the variables while the nodes of a TSP graph are monomials. Therefore, while the goal of CSP is to break the entire variable $\mvx$ into smaller cliques of variables, the goal of TSP is to break the dense monomial basis $[\mvx(I_l)]_d$ into smaller cliques of monomials.

Combining CS and TS, we can further decompose the PSD constraints in the sparse Moment-SOS hierarchy~\eqref{eq:gs:cs-moment} into smaller ones. Due to space constraints, we defer a formal presentation to Appendix~\ref{app:sec:cs-ts-relax}. The high-level intuition, however, is that the binary masks obtained from the TSP allow us to only focus on the entries of the matrix variables in~\eqref{eq:gs:cs-moment} corresponding to nonzero entries in the masks.

\subsection{Sparse Polynomial Optimization Toolbox (SPOT)}
\label{sec:gs:spot}

\begin{figure*}[htbp]
    \centering
    \begin{minipage}{\linewidth}
        \centering
        \begin{tabular}{c}
            \begin{minipage}{0.95\linewidth}
                \centering
                \includegraphics[width=\linewidth]{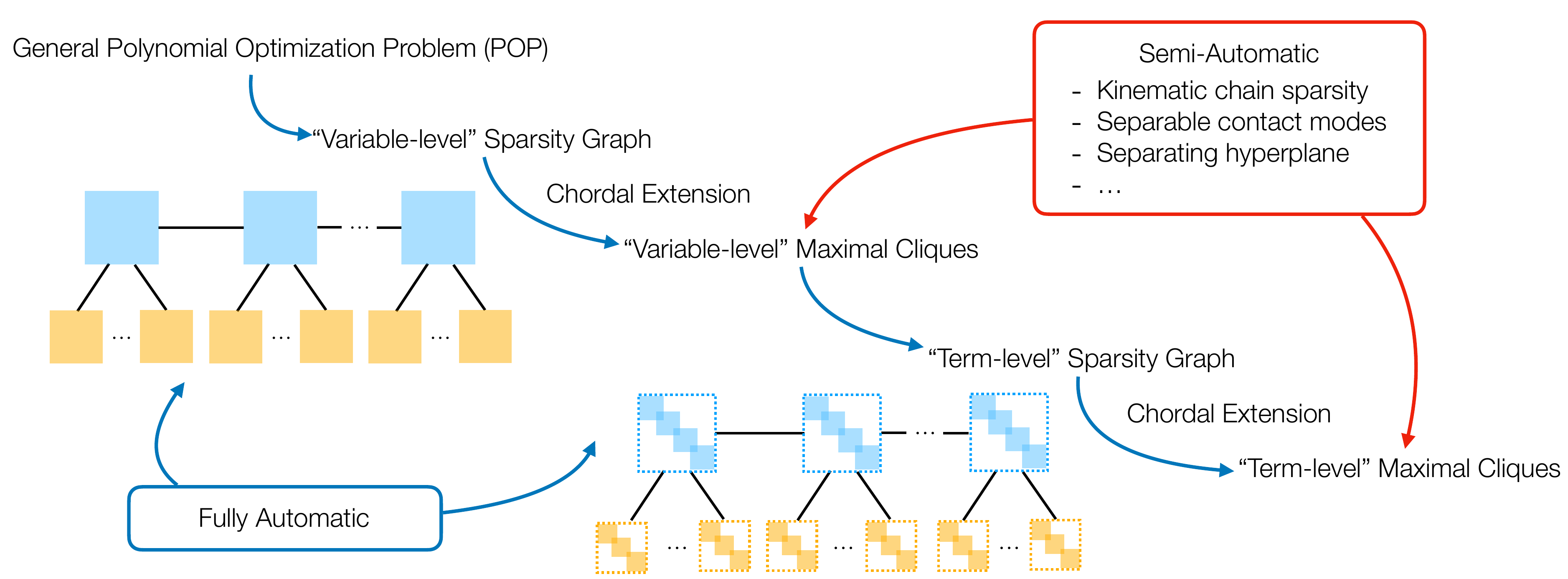}
            \end{minipage}
        \end{tabular}
    \end{minipage}

    \caption{\ksc{Overall pipeline of the Sparse Polynomial Optimization Toolbox (SPOT).
    Blue curves: the automatic detection of correlative and term sparsity patterns. Red curves: semi-automatic injection of robotics-specific sparsity patterns.} \label{fig:exp:pipeline}}
\end{figure*}

To automate the aforementioned sparsity exploitation, we develop a new C++ package \spot. Compared with the Julia package \tssos, \spot is faster and also provides more options. (i) \spot provides both \ksc{general} moment relaxation and SOS relaxation, which complements \tssos (only provides SOS relaxation). 
(ii) For both CS and TS, \spot provides four options: (a) maximal chordal extension (MAX); (b) minimal degree chordal extension (MD); (c) minimum fill chordal extension (MF); (d) user-defined. 
(iii) \spot provides a special option ``\emph{partial term sparsity}'', which enables TS only for the moment matrices and the localizing matrices (\ie inequality constraints), while applying CS to equality constraints. This heuristic approach is motivated by the observation that in SOS relaxation, tightening equality constraints merely introduces more free variables and does not significantly impact the solution time of an SDP solver. In many cases, partial TS yields tighter lower bounds while keeping the computation time nearly unchanged.
(iv) Through our Matlab and Python interface, \spot provides a way for the user to visualize the CSP and TSP graphs, as shown in Fig.~\ref{fig:demos}. \ksc{The automatic sparsity detection pipleline is illustrated in Fig.~\ref{fig:exp:pipeline}.} 

%!TEX root = ../../main.tex

% \clearpage
% \input{sections/general_sparsity.tex}
%!TEX root = ../main.tex

\section{Robotics-Specific Sparsity}
\label{sec:robotics-specific}

While automatic correlative and term sparsity (CS-TS) exploitation is powerful, it has two notable limitations. 
(a) It occasionally fails to capture time, spatial, and kino-dynamical sparsity inherent in contact-rich planning problems, such as the Markov property described in~\cite{kang2024wafr-strom}.
(b) The approach introduced in \S\ref{sec:general-sparsity-ksc} heavily depends on approximate minimal chordal extensions. Although theoretically rigorous, chordal extensions can substantially increase the size of variable or term cliques, resulting in scalability challenges.

To address these, we propose several robotics-specific sparsity patterns as auxiliary tools to complement the automatic sparsity. Similarly, we categorize robotics-specific sparsity patterns into two levels, (a) variable level and (b) term level. \ksc{The semi-automatic robotics-specific sparsity pattern injection is shown in Fig.~\ref{fig:exp:pipeline}, marked in red.}

%!TEX root = ../../main.tex

\begin{figure}[htbp]
    \centering
    \includegraphics[width=\columnwidth]{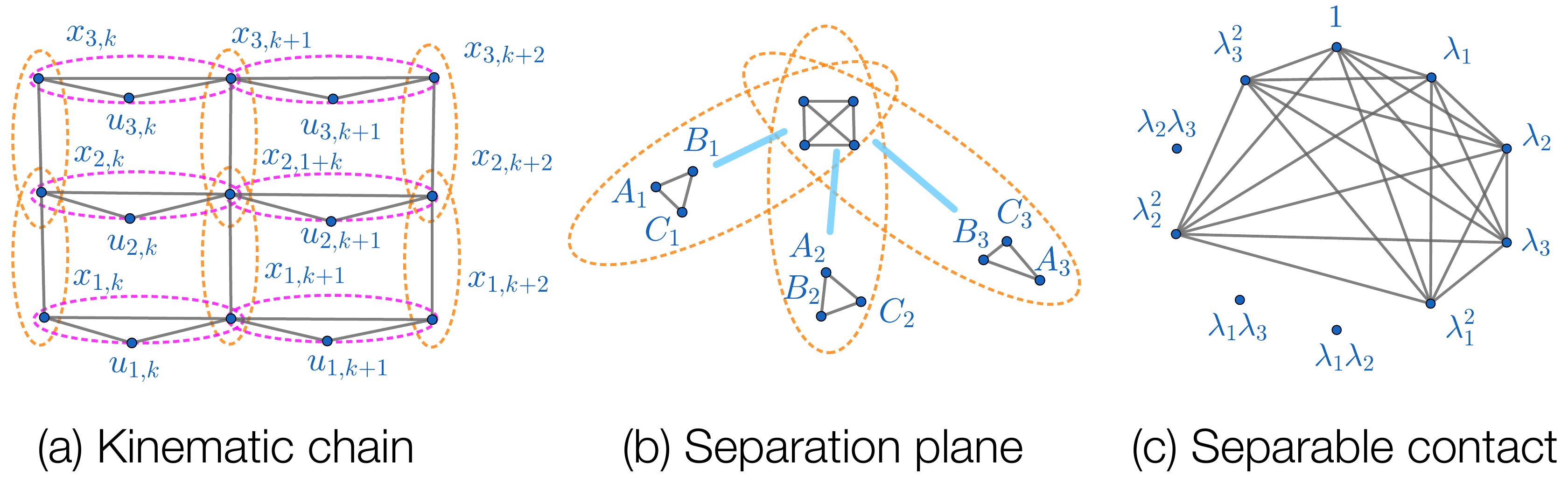}
    \caption{Robotics-specific sparsity patterns.
    \label{fig:rs:robotics-specific}}
    \vspace{-6mm}
\end{figure}

\subsection{Variable-level Robotics-Specific Sparsity}

\textbf{Kinematic chain.} Numerous robotic systems exhibit chain-like (or tree-like) mechanical structures, commonly found in manipulation and locomotion. These structures lead to a natural separation of kinematic and dynamic variables across different links. For example, consider the following system:
\begin{subequations} \label{eq:rs:kinematic-chain}
    \begin{align}
& x_{1,k+1} = x_{1,k} + u_{1,k}, \\
 &x_{2,k+1} = x_{2,k} + u_{2,k} , \\
 & x_{3,k+1} = x_{3,k} + u_{3,k} ,\\
        & (x_{1,k} - x_{2,k})^2 = r^2 , \quad (x_{2,k} - x_{3,k})^2 = r^2 \label{eq:rs:kinematic-chain:constraint}
    \end{align} 
\end{subequations}
which can be viewed as a 1-D two-link chain, with three states $(x_1, x_2, x_3)$ and two geometric constraints in~\eqref{eq:rs:kinematic-chain:constraint}. If only exploring the Markov property in~\cite{kang2024wafr-strom}, a chain of cliques will be derived, with the $k$'th clique containing 9 variables:
\begin{align}
    \left\{ x_{i,k}, x_{i,k+1}, u_{i,k} \right\}_{i=1}^3.
\end{align}
However, since $(x_1, x_2)$, $(x_2, x_3)$ forms two links, and each $u_i, i \in \left\{ 1,2,3 \right\}$ only has direct effect on $x_i$, we can further decompose the clique as three cliques of size 3:
\begin{align}
    \left\{ 
        x_{i,k}, x_{i,k+1}, u_{i,k}
    \right\}, \ i = 1,2,3
\end{align} 
and four cliques of size 2: 
\begin{align}
    \left\{ 
        x_{i,m}, x_{i+1,m}
     \right\}, \ i = 1,2, \ m = k, k+1,
\end{align}
as illustrated in Fig.~\ref{fig:rs:robotics-specific}(a). 
% If second-order relaxation is used, the former large clique will involve one positive semidefinite (PSD) cone of size $55 \times 55$, while the latter only contains three PSD cones of size $10 \times 10$ and four PSD cone of size $6 \times 6$. 
Note that this clique partition cannot be discovered by the general sparsity pattern introduced in \S\ref{sec:general-sparsity-ksc}, since the graph in Fig.~\ref{fig:rs:robotics-specific}(a) is not chordal. 

\textbf{Separation plane.} Consider a general obstacle avoidance task: both the robot and the obstacles can be decomposed as a union of convex sets. We denote the decomposition of robot as $\left\{ P_i \right\}_{i \in \enum{m}}$ and the decomposition of obstacles as $\left\{ Q_j \right\}_{j \in \enum{n}}$. For each $i$ and $j$, $P_i$ has no collision with $Q_j$ if and only if there exists a plane $H_{i,j}$ separating $P_i$ and $Q_j$. Consider $P_i$ and $Q_j$ being both 2-D polygons:
\begin{align}
    \label{eq:rs:separation-plane}
    \vspace{-10mm} A_{i,j}  v_{r,x} + B_{i,j}  v_{r,y} + C_{i,j} \ge 0, \ \forall v_r \in P_i's \text{ vertices} \\
   \vspace{-10mm} A_{i,j} v_{o,x} + B_{i,j} v_{o,y} + C_{i,j} \le 0, \ \forall v_o \in Q_j's \text{ vertices}
\end{align}
where $(A_{i,j}, B_{i,j}, C_{i,j})$ determines a 2-D separation plane (line) $H_{i,j}$. 
% Note that this formulation can be viewed as a "dual" formulation of GCS~\cite{marcucci2023sr-motion-planning-gcs}, whose decomposition happens in the free space.
For constraints~\eqref{eq:rs:separation-plane}, $(A_{i,j}, B_{i,j}, C_{i,j})$ has no direct relationship between each other. Thus, in each time step, instead of defining a large variable clique:
\begin{align}
    \label{eq:rs:separation-plane-large}
    \left\{ \text{other variables}, \left\{ (A_{i,j}, B_{i,j}, C_{i,j}) \right\}_{i \in \enum{m}, j \in \enum{n}} \right\}
\end{align}
we define $mn$ smaller variable cliques:
\begin{align}
    \label{eq:rs:separation-plane-small}
    \left\{ 
        \text{other variables}, (A_{i,j}, B_{i,j}, C_{i,j})
     \right\}_{i \in \enum{m}, j \in \enum{n}}
\end{align}
The resulting clique size is invariant to $m$ and $n$, as illustrated in Fig.~\ref{fig:rs:robotics-specific}(b). 
% In terms of the PSD cone size in the second order relaxation, the former~\eqref{eq:rs:separation-plane-large} grows $\calO(m^2 n^2)$, while the latter~\eqref{eq:rs:separation-plane-small} only contains $mn$ smaller PSD cone of size $\calO(1)$. 
What's more, one can check that the decomposition~\eqref{eq:rs:separation-plane-small} still satisfies the running intersection property (RIP) required in correlative sparsity pattern~\cite{lasserre2006msc-correlativesparse}.

\subsection{Term-level Robotics-Specific Sparsity}

\textbf{Separable contact modes.} Frequently in contact-rich planning, we will have to ``select one out of a bunch of modes'' (\cf \S\ref{app:pd:push-box} and\S\ref{app:pd:push-T-block}). It can be modelled as polynomial equalities:
\begin{align}
    \label{eq:rs:separable-contact-modes}
    & h_0 \triangleq \sum_{i \in \enum{n}} \lambda_i^2 - 1 = 0 \\
    & h_i \triangleq \lambda_i \cdot (1 - \lambda_i) = 0, \ \forall i \in \enum{n},
\end{align}
where $\lambda_i$ is a binary variable corresponding to whether the $i$-th contact mode is selected.
In variable level, there is no sparsity in~\eqref{eq:rs:separable-contact-modes}, since $\sum_{i \in \enum{n}} \lambda_i^2 = 1$ groups all $\lambda_i$'s together. However, we show that the sparsity is still rich in the term level. Consider the generation procedure of $\calA$ in term sparsity~\eqref{eq:gs:ts-A}, for the polynomial equality constraint system~\eqref{eq:rs:separable-contact-modes}:
\begin{align}
    \label{eq:rs:separable-contact-A}
    \calA = \left\{ 
        1,
        \left\{ \lambda_i \right\}_{i \in \enum{n}}, 
        \left\{ \lambda_i^2 \right\}_{i \in \enum{n}}
     \right\}.
\end{align}
Consider the special case $n = 3$ in second-order relaxation $d = 2$, and there is only one variable clique (\ie $l = 1$). By definition of the polynomial multiplier $H_2(h_i, I_1), i \in \left\{ 0,1,2,3 \right\}$ for equality constraints~\eqref{eq:gs:H}:
\begin{align}
    \hspace{-4mm} H_2(h_i, I_1)\! =\! \left[ 
        1,\lam{1},\lam{2},\lam{3},\lam{1}^2,\lam{1}\lam{2},\lam{1}\lam{3},\lam{2}^2,\lam{2}\lam{3},\lam{3}^2
     \right]\tran\!\!.\!\!\!\! 
\end{align}
If we proceed with the support extension procedure for equality constraints~\eqref{eq:gs:ts-equality-support-extension}, then we have 
\begin{align}
    B_{2,1,0}^h = & \left[ 1,1,1,1,1,0,0,1,0,1 \right] \\
    B_{2,1,1}^h = & \left[ 1,1,0,0,0,0,0,0,0,0 \right] \\
    B_{2,1,2}^h = & \left[ 1,0,1,0,0,0,0,0,0,0 \right] \\
    B_{2,1,2}^h = & \left[ 1,0,0,1,0,0,0,0,0,0 \right]
\end{align}
In the second-order moment matrix, the unmasked terms are:
\begin{equation}
    \begin{split}
        1, \left\{ \lambda_i \right\}_{i \in \enum{3}}, \left\{ \lambda_i^2 \right\}_{i \in \enum{3}}, \left\{ \lambda_i^3 \right\}_{i \in \enum{3}}, \\ \left\{ \lambda_i^4 \right\}_{i \in \enum{3}}, \left\{ \lambda_i \lambda_j^2 \right\}_{i \in \enum{3}, j \in \enum{3}, i \ne j}
       \end{split}
\end{equation}
leading to the moment matrix generated by a reduced basis:
\begin{align}
    \label{eq:rs:separable-contact-basis}
    \left\{ 1,\lam{1},\lam{2},\lam{3},\lam{1}^2,\lam{2}^2,\lam{3}^2 \right\}
\end{align}
as shown in Fig.~\ref{fig:rs:robotics-specific}(c). The key observation is that there is no $\lam{i}\lam{j} (i \ne j)$ in the new basis (i.e., the contact modes are separated).
For a general $n$, the reduced basis is of size $2n+1$, which is much smaller than the standard monomial basis of size $\frac{(n+1)(n+2)}{2}$. 

\textbf{Separable contact forces.} In contact-implicit formulation, each possible contact is modelled as a set of complementary constraints. Suppose there are $n$ contact points, then a typical dynamics formulation is (\cf \S\ref{app:pd:push-bot} and \S\ref{app:pd:planar-hand}):
\begin{align}
    \label{eq:rs:contact-implicit}
    & 0 \le \lam{i} \perp g_i(x) \ge 0, \ i \in \enum{n} \\
    & \sum_{i \in \enum{n}} f_i(\lambda_i, x, u) = 0
\end{align}
where $x \in \Real{n_x}$ is the system state and $u \in \Real{n_u}$ is the control input. Similar to the contact mode case~\eqref{eq:rs:separable-contact-modes}, $\sum_{i \in \enum{n}} f_i(\lambda_i, x, u)$ groups all $\lambda_i$'s together. Thus, there is no variable-level sparsity pattern. However, inspired by the reduced basis introduced in~\eqref{eq:rs:separable-contact-basis}, we can directly write down an extended reduced basis containing $x$ and $u$. For simplicity, assume $x$ and $u$ are both of one dimension, $f_i$ is of quadratic form and linear in $\lambda_i$, then the reduced basis is
\begin{align}
    \left\{ \substack{
        1, x, u, x^2, xu, u^2, \\ \left\{ \lambda_i \right\}_{i \in \enum{n}}, 
        \left\{ \lambda_i^2 \right\}_{i \in \enum{n}}, \left\{ x \lambda_i \right\}_{i \in \enum{n}}, \left\{ u \lambda_i \right\}_{i \in \enum{n}}}
     \right\}
\end{align}
which grows linearly in $n$. 

\ksc{Although we introduce these robotics-specific sparsity patterns in the context of contact-rich motion planning, many—such as kinematic-chain structures and supporting-hyperplane constraints—are generic and apply equally well to tasks that do not involve contact.} 

% \input{sections/roboust-extraction.tex}
%!TEX root = ../main.tex

\section{Experiments}
\label{sec:exp}

%!TEX root = ../../main.tex

\begin{figure}[htbp]
    \centering
    \begin{minipage}{\columnwidth}
        \centering
        \begin{tabular}{cc}
            \begin{minipage}{0.35\columnwidth}
                \centering
                \includegraphics[width=\columnwidth]{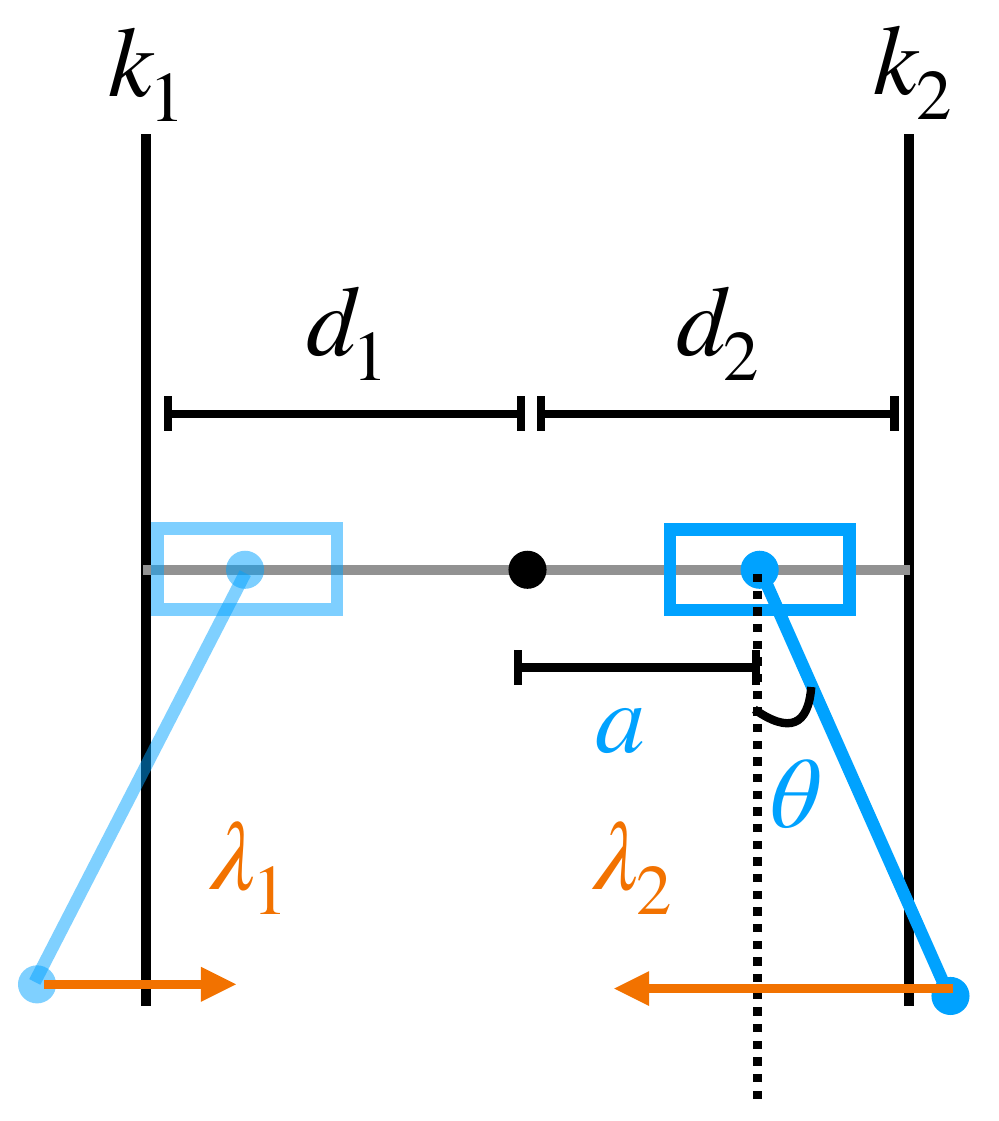}
                (a) 
            \end{minipage}
            
            \begin{minipage}{0.50\columnwidth}
                \centering
                \includegraphics[width=\columnwidth]{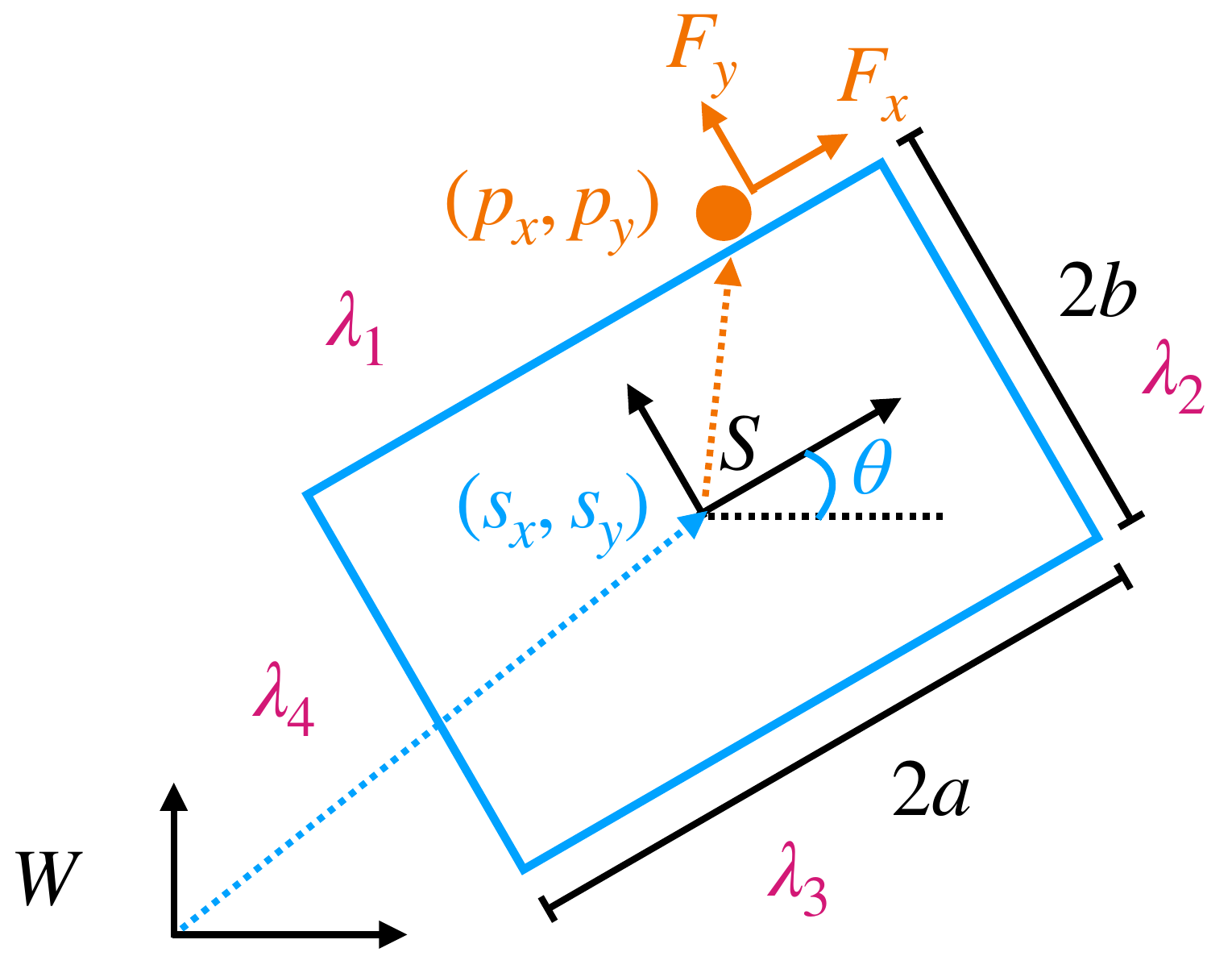}
                (b) 
            \end{minipage}
        \end{tabular}
    \end{minipage}

    \begin{minipage}{\columnwidth}
        \begin{tabular}{cc}
            \begin{minipage}{0.52\columnwidth}
                \centering
                \includegraphics[width=\columnwidth]{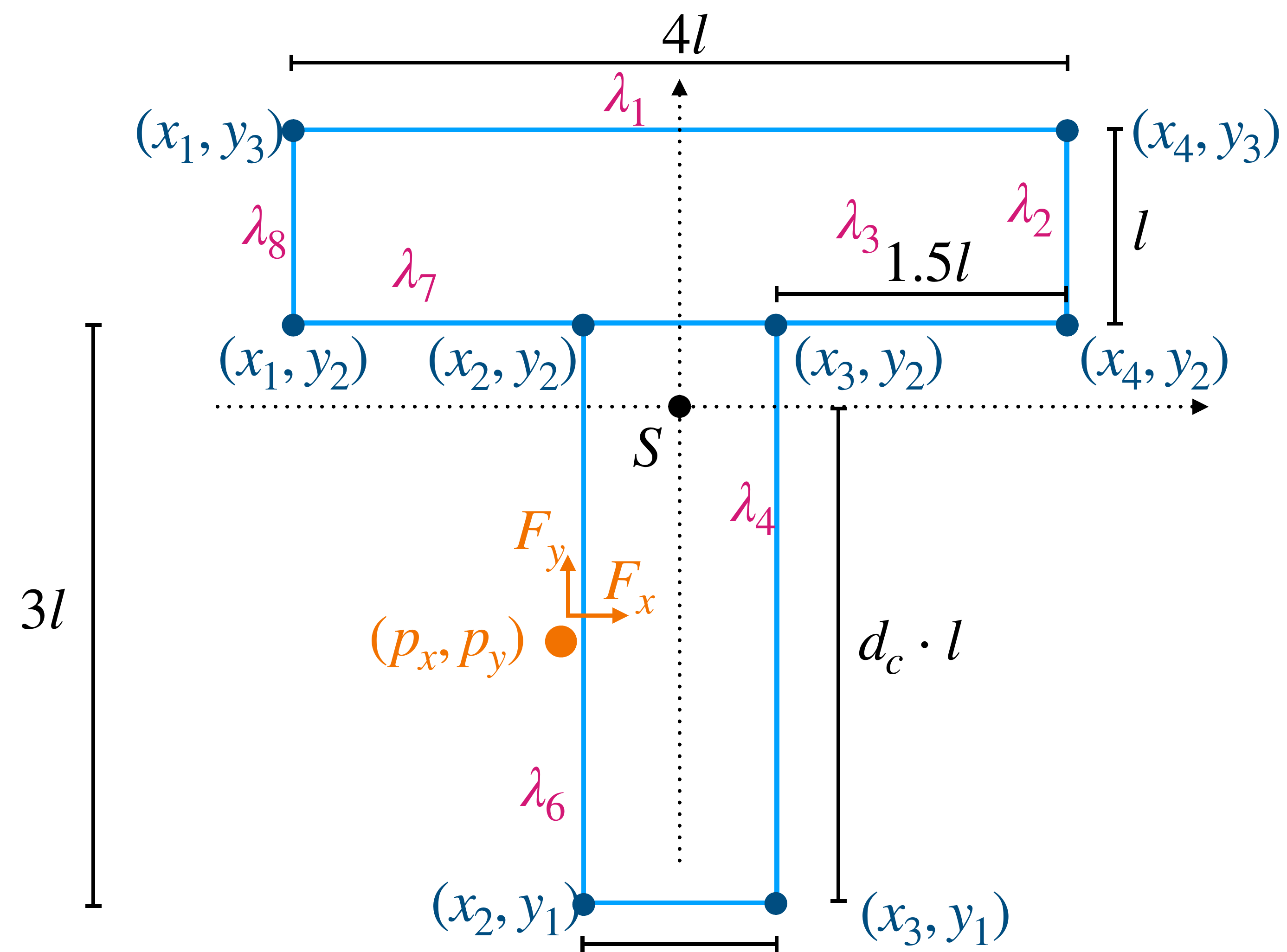}
                (c) 
            \end{minipage}

            \begin{minipage}{0.35\columnwidth}
                \centering
                \includegraphics[width=\columnwidth]{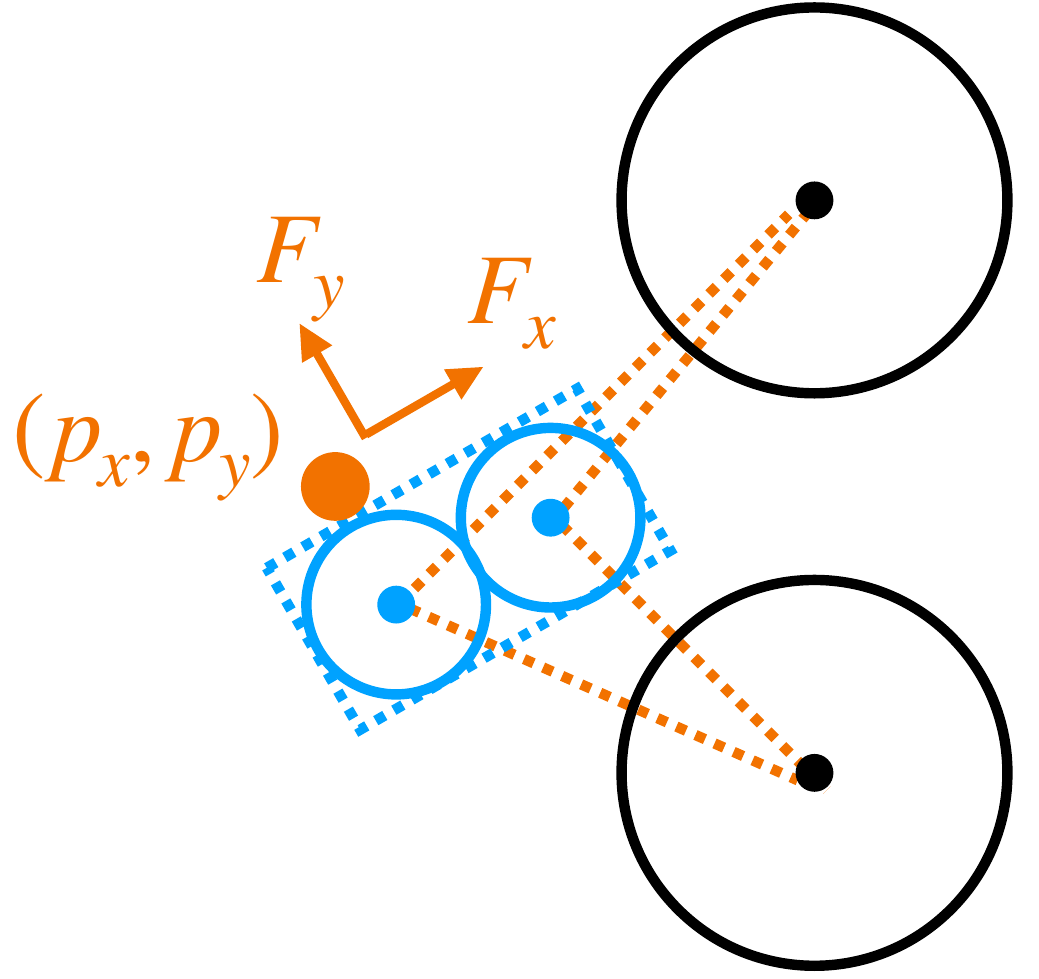}
                (d) 
            \end{minipage}
        \end{tabular}
    \end{minipage}

    \vspace{2mm}
    \begin{minipage}{\columnwidth}
        \begin{tabular}{c}
            \begin{minipage}{0.90\columnwidth}
                \centering
                \includegraphics[width=\columnwidth]{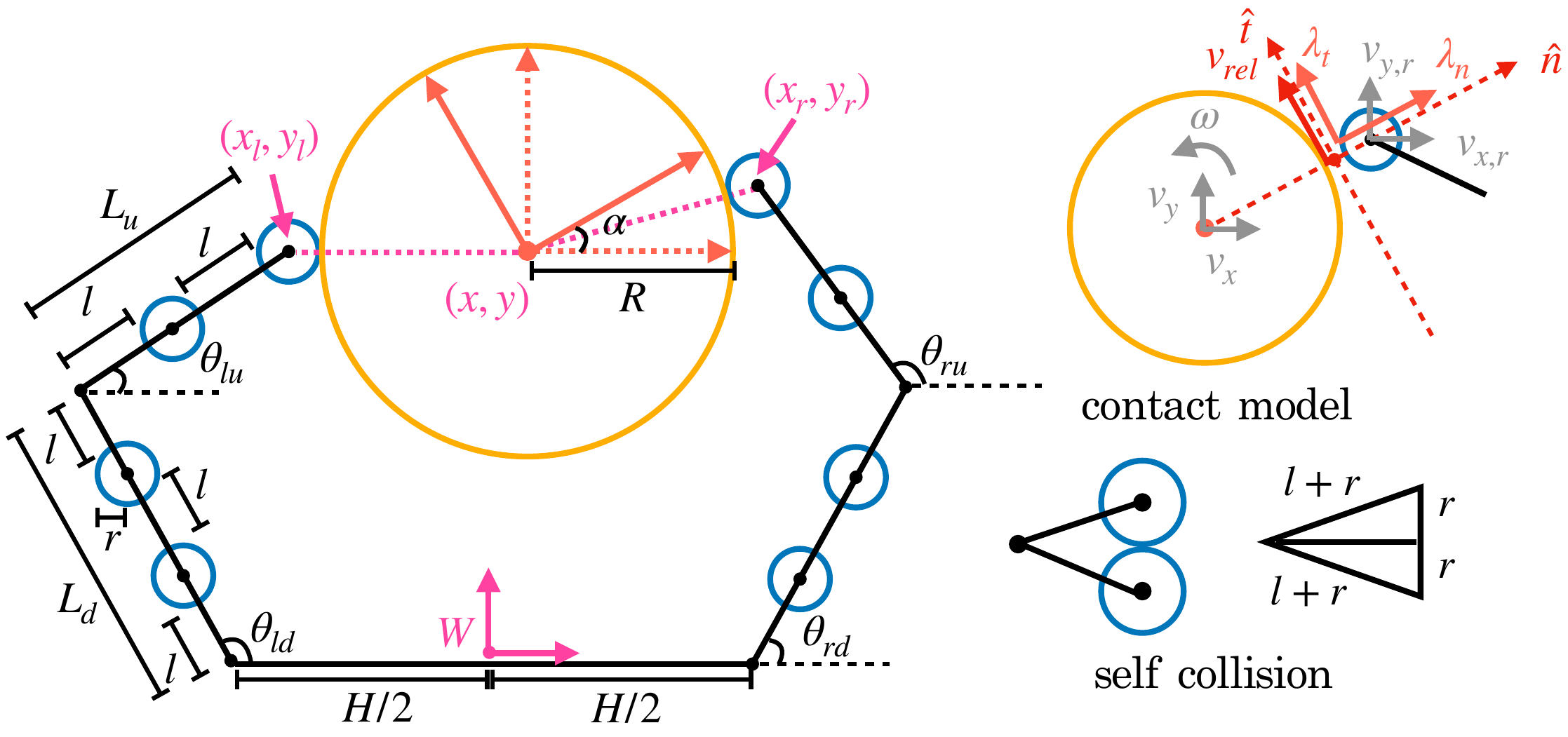}
                (e) 
            \end{minipage}
        \end{tabular}
    \end{minipage}

    \caption{illustrations of five contact-rich planning tasks. (a) Push Bot; (b) Push Box; (c) Push T-block; (d) Push Box with a Tunnel; (e) Planar Hand. \label{fig:exp:illustration}}
\end{figure}

\textbf{General setup.} We consider five contact-rich problems.  
\begin{enumerate}
    \item \textbf{Push Bot:} A cart-pole system between two soft walls, shown in Fig.~\ref{fig:exp:illustration}(a). The objective is to stabilize the cart-pole at $(a, \theta) = (0, \pi)$. For polynomial dynamics and other detailed settings, please refer to Appendix~\ref{app:pd:push-bot}.  

    \item \textbf{Push Box:} A simple pusher-slider system, illustrated in Fig.~\ref{fig:exp:illustration}(b). The goal is to push the box from an initial configuration $(\sx, \sy, \theta)$ to a target configuration. There are 4 possible contact modes at each time step. Detailed settings can be found in Appendix~\ref{app:pd:push-box}.  

    \item \textbf{Push T-Block:} Similar to Push Box, but the box is replaced with a T-block, resulting in 8 possible contact modes at each time step, as shown in Fig.~\ref{fig:exp:illustration}(c). Please refer to Appendix~\ref{app:pd:push-T-block} for detailed settings.  

    \item \textbf{Push Box with a Tunnel:} Similar to Push Box, except that two circular obstacles form a tunnel along the box's path to its goal, as shown in Fig.~\ref{fig:exp:illustration}(d). Detailed settings are provided in Appendix~\ref{app:pd:push-box-tunnel}.  

    \item \textbf{Planar Hand:} A two-fingered system rotating a 2D disk in a horizontal plane, as shown in Fig.~\ref{fig:exp:illustration}(e). The goal is to rotate the disk by $360^\circ$ using two fingertip contacts while minimizing the translation of the disk's center of mass. See Appendix~\ref{app:pd:planar-hand} for further details.  
\end{enumerate}  

Additionally, extensive real-world validations are conducted for the Push T task. Experiments were performed on a high-performance workstation equipped with a 2.7 GHz AMD 64-Core sWRX8 Processor and 1 TB of RAM, enabling \MOSEK~\cite{aps2019ugrm-mosek-sdpsolver} to solve large-scale problems utilizing 64 threads.

\textbf{Conversion speed.} 
We compare \spot and \tssos in terms of conversion time across the five planning problems, considering different CS and TS options. In all cases, the planning horizon is set to $N = 30$. For the Push Box with a Tunnel problem, the CS relaxation order is set to $d = 3$ to obtain a tighter lower bound, while for the other examples, $d$ is set to $2$.
The results are summarized in Table~\ref{tab:performance_comparison}. \spot consistently outperforms \tssos, achieving at least a $2\times$ speedup. For large-scale problems such as Push Box with a Tunnel, \spot achieves approximately a $5\times$ speedup.
For the Planar Hand problem, the automatic CS pattern generation produces clique sizes exceeding 20 through both MF and MD methods, resulting in a large-scale SDP with over $1$ million constraints. Since \tssos tightly integrates its conversion and SDP solving processes, it is difficult to isolate the conversion time. Therefore, we only report \spot's conversion time for this case, which remains under 100 seconds despite the problem's scale.
\begin{table}[htbp]
    \centering
    \begin{adjustbox}{width=\linewidth}
    \begin{tabular}{|c|c|ccc|ccc|}
        \hline
        \textbf{Sparsity}& \textbf{CS} 
        & \multicolumn{3}{c|}{\textbf{MF}} 
        & \multicolumn{3}{c|}{\textbf{MD}} \\ \cline{2-8}
        \textbf{Examples} & \textbf{TS}
        & \textbf{MAX} & \textbf{MF} & \textbf{MD} 
        & \textbf{MAX} & \textbf{MF} & \textbf{MD} \\ \hline

        \multirow{3}{*}{Push Bot} 
        & MOMENT & $2.33$ & $2.29$ & $2.13$ & $2.03$ & $1.99$ & $1.82$ \\ 
        & SOS    & $2.39$ & $2.43$ & $2.26$ & $2.09$ & $2.13$ & $1.97$ \\ 
        & TSSOS  & $4.32$ & $14.95$ & $19.58$ & $3.49$ & $15.90$ & $16.91$ \\ \hline

        \multirow{3}{*}{Push Box} 
        & MOMENT & $1.68$ & $1.58$ & $1.52$ & $1.39$ & $1.31$ & $1.24$ \\ 
        & SOS    & $1.64$ & $1.65$ & $1.57$ & $1.37$ & $1.37$ & $1.30$ \\ 
        & TSSOS  & $4.56$ & $8.83$ & $10.12$ & $3.94$ & $9.60$ & $8.92$ \\ \hline

        \multirow{3}{*}{Push T} 
        & MOMENT & $3.67$ & $3.58$ & $3.29$ & $4.01$ & $3.95$ & $3.48$ \\ 
        & SOS    & $3.70$ & $3.77$ & $3.47$ & $4.06$ & $4.16$ & $3.67$ \\ 
        & TSSOS  & $15.54$ & $42.86$ & $43.41$ & $13.13$ & $43.22$ & $45.02$ \\ \hline

        \multirow{3}{*}{Tunnel} 
        & MOMENT & $21.19$ & $18.03$ & $16.19$ & $19.78$ & $19.79$ & $17.69$ \\ 
        & SOS    & $20.55$ & $18.62$ & $16.81$ & $22.76$ & $20.81$ & $18.63$ \\ 
        & TSSOS  & $90.37$ & $109.25$ & $112.20$ & $83.47$ & $129.17$ & $125.63$ \\ \hline

        \multirow{3}{*}{Planar Hand} 
        & MOMENT & $74.35$ & $75.60$ & $64.98$ & $67.54$ & $68.31$ & $58.53$ \\ 
        & SOS    & $88.60$ & $91.35$ & $80.70$ & $80.70$ & $83.14$ & $70.75$ \\ 
        & TSSOS  & -- & -- & -- & -- & -- & -- \\ \hline
    \end{tabular}
    \end{adjustbox}
    \caption{Conversion time comparison between \spot and \tssos across examples, with different CS-TS options. \label{tab:performance_comparison}}
\end{table}

\textbf{Self-defined variable cliques.} 
Since the automatic sparsity exploitation mechanism may fail to detect robotics-specific sparsity, we adopt the following clique generation procedure:
\begin{enumerate}
    \item Generate a general sparsity pattern using \spot.
    \item Manually inspect the variable cliques to determine whether certain robotics-specific variable-level sparsity patterns can be incorporated.
    \item Modify the generated cliques and resend them to \spot using the ``SELF'' option.
\end{enumerate}
    
    For example, in the Planar Hand task, leveraging the kinematic chain pattern discussed in \S\ref{sec:robotics-specific}, we manually partition the variables at each time step into 14 smaller cliques, with sizes ranging from 6 to 14. Due to space constraints, these cliques are detailed in Appendix~\ref{app:sec:self-cliques}. These manually defined cliques precisely correspond to the red circles in Fig.~\ref{fig:demos}(a), illustrating the specific sparsity pattern.

\textbf{Robust minimizer extraction.} Due to the inherent complexity of contact-rich planning problems, it is not uncommon to encounter two types of relaxation ``failure cases'':  
\begin{enumerate}
    \item The sparse Moment-SOS Hierarchy is not tight.  
    \item The sparse Moment-SOS Hierarchy is tight but admits multiple solutions.  
\end{enumerate}  
In both cases, the moment matrices fail to attain rank 1 (or other general tightness certificates).  
To extract minimizers, \tssos and~\cite{kang2024wafr-strom} both employ the same ``naive'' approach: obtain the degree-1 submatrix of each moment matrix, then average the normalized eigenvectors across different variable cliques. While straightforward to implement, this method is not robust in practice, often leading to infeasible local rounding and large suboptimality gaps.  
On the other hand,~\cite{klep2018siopt-minimizer-extraction-robust} demonstrates that the Gelfand-Naimark-Segal (GNS) construction provides a robust approach for minimizer extraction from a single moment matrix. Inspired by this, we propose the following heuristic algorithm for minimizer extraction in the presence of multiple moment matrices:  
\begin{enumerate}
    \item For each moment matrix, extract minimizers along with their associated weights using the GNS construction.  
    \item Select the minimizer with the highest weight and average it across different variable cliques.  
\end{enumerate}  
This seemingly minor modification significantly improves robustness compared to the ``naive'' extraction method. We implemented the new minimizer extraction scheme in our \spot package. A detailed discussion of GNS is beyond the scope of this paper; we refer interested readers to~\cite{klep2018siopt-minimizer-extraction-robust}. 

\begin{table*}[t]
    \centering
    \begin{adjustbox}{width=0.95\linewidth}
    \begin{tabular}{|c|ccc|ccc|ccc|}
    \hline
    \multirow{1}{*}{\textbf{Examples}} 
               & \multicolumn{3}{c|}{SELF + NON}                             
               & \multicolumn{3}{c|}{SELF + MAX}                             
               & \multicolumn{3}{c|}{SELF + MF}                              \\ \cline{2-10} 
               & $\eta_g$ (\%)      & $-\log_{10}(\eta_\kkt)$      & time (s)        
               & $\eta_g$ (\%)    & $-\log_{10}(\eta_\kkt)$     & time (s)        
               & $\eta_g$ (\%)    & $-\log_{10}(\eta_\kkt)$     & time (s)        \\ \hline
    Push Bot    & $0.08$   & $5.90$        & $22.07$           
               & $9.28$   & $3.67$        & $14.01$           
               & $32.39$  & $5.47$        & $5.21$            \\ \hline
    Push Box    & $0.15$   & $5.65$        & $8.90$                
               & $0.20$   & $5.70$        & $5.76$                 
               & $13.77$  & $6.37$        & $1.67$                 \\ \hline
    Push T      & $7.83$   & $5.30$        & $46.20$           
               & $17.10$  & $5.67$        & $29.35$           
               & $35.98$  & $5.50$        & $4.99$            \\ \hline
    Tunnel    & $4.89$   & $5.52$        & $342.31$          
               & $5.22$   & $5.35$        & $266.17$          
               & $8.20$   & $5.04$        & $33.31$           \\ \hline
    Planar Hand & $22.26$  & $5.00$        & $343.79$                 
               & $23.97$  & $5.15$        & $143.42$                
               & $25.61$  & $3.94$        & $54.00$                 \\ \hline
    \end{tabular}
    \end{adjustbox}
    \caption{Comparison of suboptimality gap $\eta_g$, max KKT residual $\eta_\kkt$, and \MOSEK solving times for different tasks under SELF + NON, SELF + MAX, and SELF + MF settings. Throughout the table, we only consider SOS relaxation. \label{tab:results}}
    
\end{table*}
\begin{table*}[t]
    \centering
    \begin{adjustbox}{width=0.95\linewidth}
    \begin{tabular}{|c|ccc|ccc|ccc|}
    \hline
    \multirow{2}{*}{\textbf{Examples}} 
               & \multicolumn{3}{c|}{SELF + NON}                             
               & \multicolumn{3}{c|}{SELF + MAX}                             
               & \multicolumn{3}{c|}{SELF + MF}                              \\ \cline{2-10} 
               & Constraint & PSD Cone    & Max PSD          
               & Constraint & PSD Cone    & Max PSD
               & Constraint & PSD Cone    & Max PSD         \\ 
               & Number     & Number      & Cone Size        
               & Number     & Number      & Cone Size           
               & Number     & Number      & Cone Size         \\ \hline
    Push Bot    & $45183$   & $1429$        & $66$           
               & $32807$   & $6310$        & $32$           
               & $11457$  & $14489$        & $11$            \\ \hline
    Push Box    & $38869$   & $1316$        & $45$                
               & $27631$   & $6043$        & $35$                 
               & $12212$  & $11456$        & $9$                 \\ \hline
    PushT      & $101080$   & $1476$        & $91$           
               & $67656$  & $8589$        & $59$           
               & $27777$  & $18875$        & $13$            \\ \hline
    Tunnel2    & $279547$   & $1260$        & $165$          
               & $222503$   & $27549$        & $12$          
               & $53553$   & $46322$        & $13$           \\ \hline
    Planar Hand & $405740$  & $4733$        & $120$                 
               & $256654$  & $25091$        & $15$                
               & $104097$  & $57569$        & $13$                 \\ \hline
    \end{tabular}
    \end{adjustbox}
    \caption{Comparison of constraint number, PSD cone number, and max PSD cone size for different tasks under SELF + NON, SELF + MAX, and SELF + MF settings. Throughout the table, only SOS relaxation are considered.}
    \label{tab:sdp-config}
\end{table*}
%!TEX root = ../../main.tex

\begin{figure*}[t!]
    \centering
    \begin{minipage}{\linewidth}
        \centering
        \begin{tabular}{c}
            \begin{minipage}{\linewidth}
                \centering
                \includegraphics[width=\linewidth]{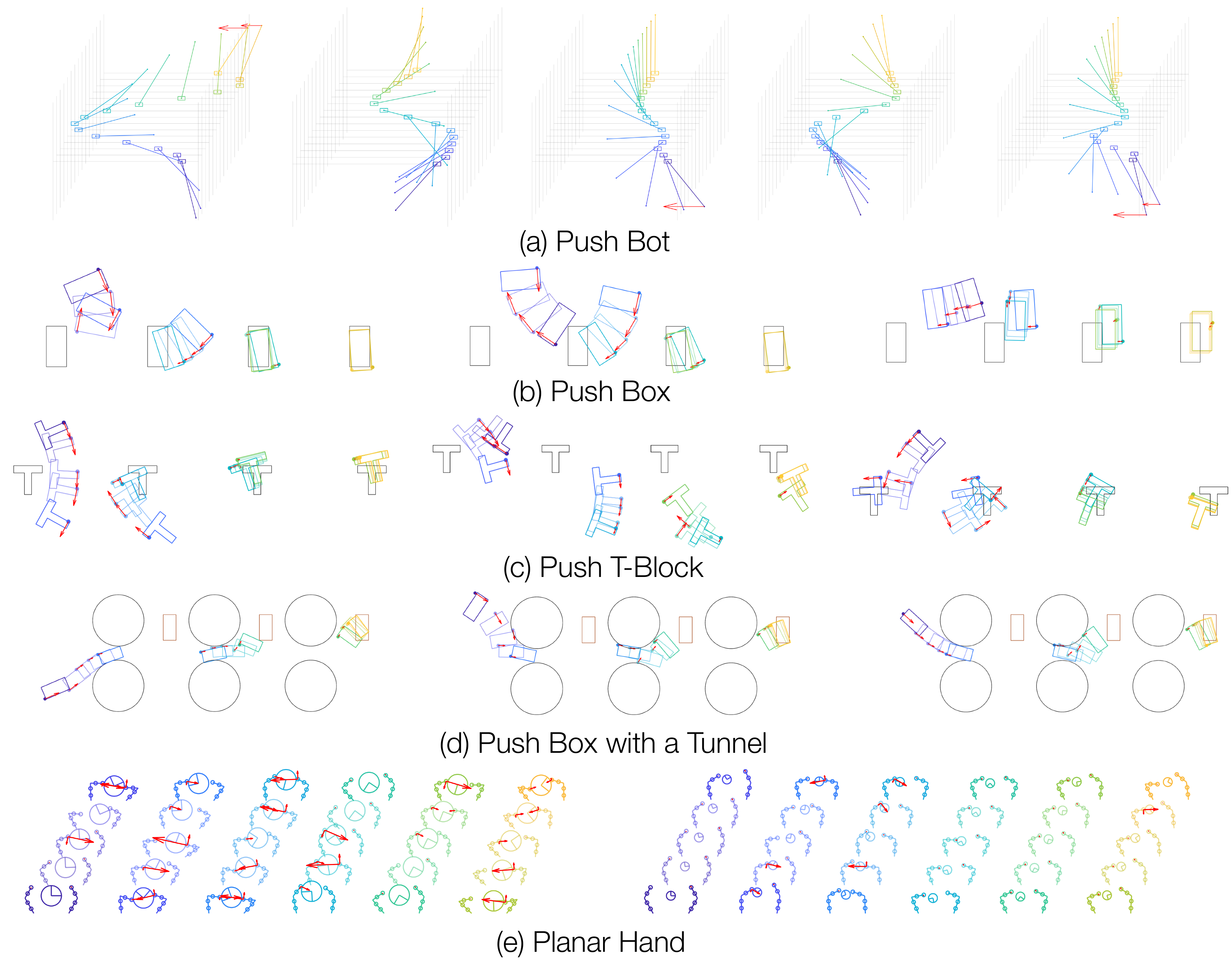}
            \end{minipage}
        \end{tabular}
    \end{minipage}

    \caption{More globally optimal trajectories from \spot. \label{fig:exp:more-simulation}}
\end{figure*}

\textbf{Numerical results.} The results are presented in Table~\ref{tab:results}. The planning horizon $N$ is set to $30$. For the Push Bot, Push Box, and Push T Block tasks, we evaluate 10 random initial states. For the Push Box with a Tunnel and Planar Hand tasks, we consider 5 random initial states. All reported statistics represent mean values.  
From semidefinite relaxation, we can get a lower bound $f_{\text{lower}}$ of original nonconvex POP. After extracting solution, we use an in-house local solver\footnote{\href{https://github.com/ComputationalRobotics/CRISP}{https://github.com/ComputationalRobotics/CRISP}}~\cite{li2025rss-crisp} to round a feasible solution with an upper bound $f_{\text{upper}}$. The suboptimality gap is defined as:
\begin{align}
    \label{eq:exp:suuboptimality-gap}
    \eta_g := \frac{
        \abs{f_{\text{lower}} - f_{\text{upper}}}
    }{
        1 + \abs{f_{\text{lower}}} + \abs{f_{\text{upper}}}
    }
\end{align}
We also report max KKT residual $\eta_\kkt$ and \MOSEK solving time.   
Since SOS relaxation tends to generate much less constraint numbers compared to moment relaxation, and \MOSEK is sensitive to constraint numbers, we only test SOS relaxations. 

The numerical results reveal a clear trade-off between computational efficiency and relaxation tightness. When TS is not enabled, all average suboptimality gaps remain below $10\%$, except for the Planar Hand task. However, solving large-scale problems can take up to 5 minutes.  
Enabling TS with MF significantly improves computational efficiency. For the Push Bot, Push Box, and Push T Block tasks, we achieve near real-time solving speeds. However, this comes at the cost of significantly larger suboptimality gaps.  
\ksc{In all of our case studies, the local solver recovered trajectories that succeed the tasks.}
Notably, there are two instances where \MOSEK fails to solve the SDP to high accuracy: (a) Push Bot, SELF + MAX; (b) Planar Hand, SELF + MF. The underlying causes of these failures remain unknown. We also report the corresponding SDP constraint number, PSD cone number, and max PSD cone size in Table~\ref{tab:sdp-config}. The demonstration of global optimal trajectories can be found in Fig.~\ref{fig:exp:more-simulation}.

%!TEX root = ../../main.tex

\begin{figure*}[t!]
    \centering
    \begin{minipage}{\linewidth}
        \centering
        \begin{tabular}{c}
            \begin{minipage}{0.99\linewidth}
                \centering
                \includegraphics[width=\linewidth]{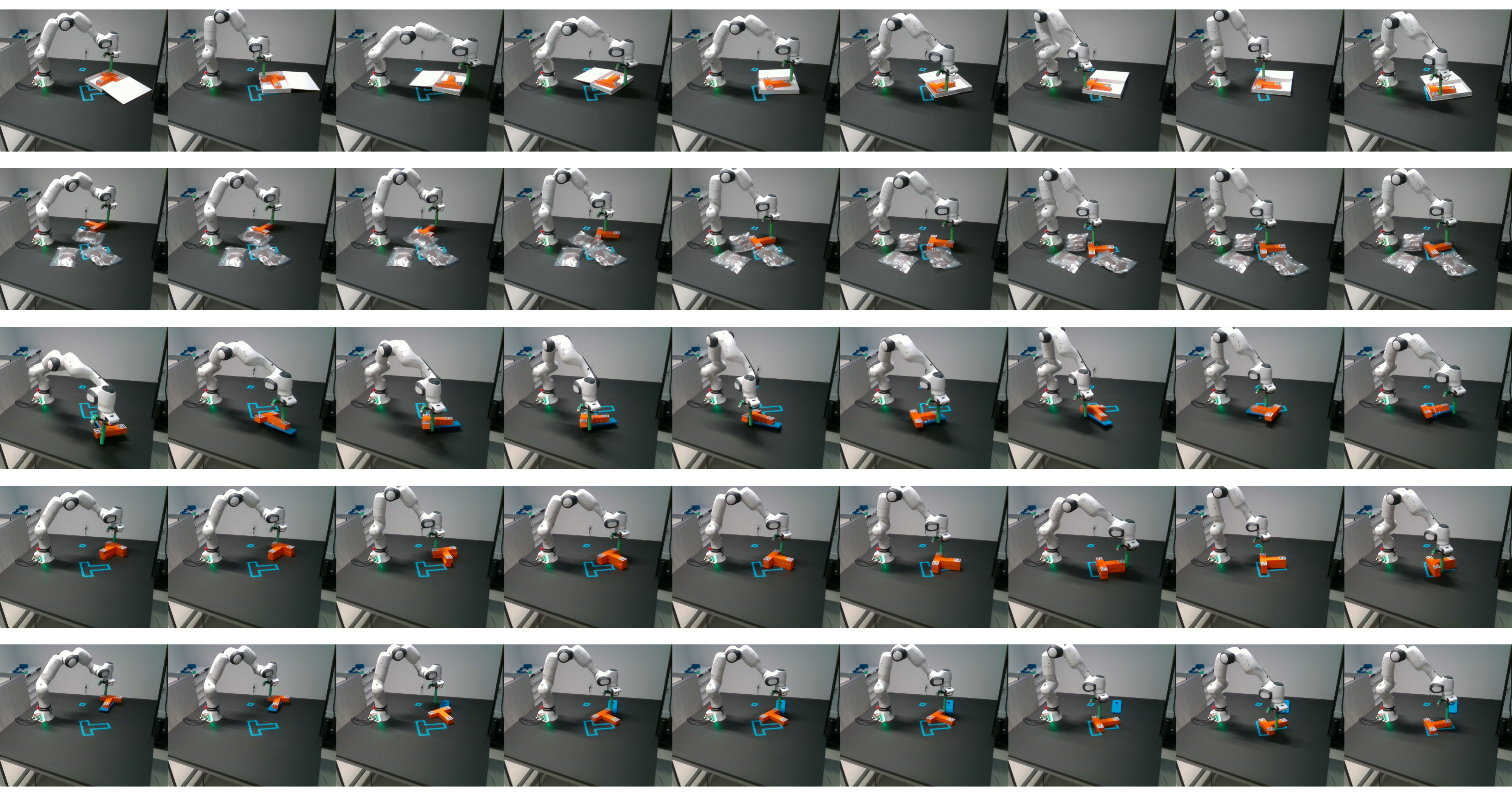}
            \end{minipage}
        \end{tabular}
    \end{minipage}

    \caption{More demonstrations from real-world ``dirty'' push-T tasks. From top to bottom: T-block (1) contained in a box; (2) sits on top of some irregular surfaces; (3) attached to some irregular objects; (4) stacked with another T-block; (5) put on some irregular objects. \label{fig:exp:more-realworld}}
\end{figure*}

\textbf{Real-world Push T-Block validations.} We extensively validate our global optimal policy in the real world Push T-Block setting, as illustrated in Fig.~\ref{fig:demos}(c). We use AprilTag~\cite{olson2011icra-apriltag} to get accurate pose estimation of the T-block. Due to the global optimality of our approach, a planning horizon of $N = 5$ is sufficient to generate high-quality planned trajectories. At each time step, we execute the pusher's first action and then re-plan. The average re-planning time is $3.7$ seconds, with CS set to MF and TS set to NON.
We evaluate our framework in 20 random trials, evenly split between two categories: clean push-T tasks, which involve standard push-T scenarios without disturbances, and ``dirty'' push-T tasks, where severe model mismatches and external disturbances are introduced to assess the planner's robustness. These disturbances include cases where the T-block is wrapped in a cable, put on some irregular surface, or contained within a small box. These three scenarios correspond to lines 2-4 in Fig.~\ref{fig:demos}(c).
Note that all these ``dirty'' push-T cases make our modeling (which is very simple) effectively ``wrong''.
However, our planner demonstrates remarkable efficiency and robustness across all test cases, achieving a $100\%$ success rate while naturally accommodating a wide range of initial states and challenging environments. 
For more demonstrations, please see Fig.~\ref{fig:exp:more-realworld} and our attached supplementary materials.

% \begin{quote}
%     \textit{All models are wrong, but some are ``powerful'', with ``global optimization''.}
% \end{quote}

% \ksc{Shucheng: do we need the last sentence? One of the reviews asked us to remove it.}

%!TEX root = main.tex

\section{Conclusion}
\label{sec:conclusion}

We introduced a new paradigm to contact-rich motion planning by exploiting both generic and robotics-specific sparsity patterns within semidefinite relaxations. Our method efficiently exploits correlative, term, and robotics-specific sparsity, enabling near-global optimization of complex robotic motion planning tasks. Through the Sparse Polynomial Optimization Toolbox (\spot), we automated sparsity detection, significantly reducing computation time while maintaining solution quality. Extensive experiments, including various real-world push-T validations, demonstrated the robustness of our approach. 

\textbf{Limitations and future work.} Despite its efficiency, our approach faces challenges in scalability, suboptimality gaps, and real-time execution. Large-scale problems with many contact modes still pose computational bottlenecks, and automatic sparsity detection can lead to oversized relaxations. Future improvements can explore GPU-accelerated first-order SDP solvers~\cite{kang2024wafr-strom}, hybrid relaxations, and learning-based heuristics to further advance real-time and large-scale applications.
\ksc{Moreover, at present, sparsity patterns are crafted individually for each problem, but we anticipate that our open-source \spot package—designed for rapid experimentation with alternative sparsity structures—will spur the discovery of many more.}

\section*{Acknowledgement}

We thank Jie Wang for the help on \tssos and sparse Moment-SOS relaxations; and Tao Pang for useful discussions about contact modeling.

% \cleaage
\bibliographystyle{plainnat}
\bibliography{refs}

\begin{thebibliography}{52}
\providecommand{\natexlab}[1]{#1}
\providecommand{\url}[1]{\texttt{#1}}
\expandafter\ifx\csname urlstyle\endcsname\relax
  \providecommand{\doi}[1]{doi: #1}\else
  \providecommand{\doi}{doi: \begingroup \urlstyle{rm}\Url}\fi

\bibitem[ApS(2019)]{aps2019ugrm-mosek-sdpsolver}
Mosek ApS.
\newblock Mosek optimization toolbox for matlab.
\newblock \emph{User's Guide and Reference Manual, Version}, 4\penalty0 (1), 2019.

\bibitem[Aydinoglu and Posa(2022)]{aydinoglu2023icra-realtime-multicontact-mpc-admm}
Alp Aydinoglu and Michael Posa.
\newblock Real-time multi-contact model predictive control via admm.
\newblock In \emph{2022 International Conference on Robotics and Automation (ICRA)}, pages 3414--3421. IEEE, 2022.

\bibitem[Aydinoglu et~al.(2021)Aydinoglu, Sieg, Preciado, and Posa]{aydinoglu2021tro-stabilization-complementary}
Alp Aydinoglu, Philip Sieg, Victor~M Preciado, and Michael Posa.
\newblock Stabilization of complementarity systems via contact-aware controllers.
\newblock \emph{IEEE Transactions on Robotics}, 38\penalty0 (3):\penalty0 1735--1754, 2021.

\bibitem[Bodlaender and Koster(2010)]{Hans2010iac-treewidth-computations}
Hans~L. Bodlaender and Arie~M.C.A. Koster.
\newblock Treewidth computations i. upper bounds.
\newblock \emph{Information and Computation}, 208\penalty0 (3):\penalty0 259--275, 2010.
\newblock ISSN 0890-5401.

\bibitem[Chatzinikolaidis and Li(2021)]{chatzinikolaidis2021ral-traopt-contact-rich-implicit-ddp}
Iordanis Chatzinikolaidis and Zhibin Li.
\newblock Trajectory optimization of contact-rich motions using implicit differential dynamic programming.
\newblock \emph{IEEE Robotics and Automation Letters}, 6\penalty0 (2):\penalty0 2626--2633, 2021.

\bibitem[Chen et~al.(2021)Chen, Culbertson, Lepert, Schwager, and Bohg]{chen2021iros-traopt-tree-search-multi-contact}
Claire Chen, Preston Culbertson, Marion Lepert, Mac Schwager, and Jeannette Bohg.
\newblock Trajectotree: Trajectory optimization meets tree search for planning multi-contact dexterous manipulation.
\newblock In \emph{2021 IEEE/RSJ International Conference on Intelligent Robots and Systems (IROS)}, pages 8262--8268. IEEE, 2021.

\bibitem[Cheng et~al.(2022)Cheng, Huang, Hou, and Mason]{cheng2022icra-contact-mode-quasidynamic}
Xianyi Cheng, Eric Huang, Yifan Hou, and Matthew~T Mason.
\newblock Contact mode guided motion planning for quasidynamic dexterous manipulation in 3d.
\newblock In \emph{2022 International Conference on Robotics and Automation (ICRA)}, pages 2730--2736. IEEE, 2022.

\bibitem[Di~Carlo et~al.(2018)Di~Carlo, Wensing, Katz, Bledt, and Kim]{di2018iros-dynamic-locomotion}
Jared Di~Carlo, Patrick~M Wensing, Benjamin Katz, Gerardo Bledt, and Sangbae Kim.
\newblock Dynamic locomotion in the mit cheetah 3 through convex model-predictive control.
\newblock In \emph{2018 IEEE/RSJ international conference on intelligent robots and systems (IROS)}, pages 1--9. IEEE, 2018.

\bibitem[Ding et~al.(2020)Ding, Li, and Park]{ding2020iros-motionplanning-multilegged-mixedinteger}
Yanran Ding, Chuanzheng Li, and Hae-Won Park.
\newblock Kinodynamic motion planning for multi-legged robot jumping via mixed-integer convex program.
\newblock In \emph{2020 IEEE/RSJ International Conference on Intelligent Robots and Systems (IROS)}, pages 3998--4005. IEEE, 2020.

\bibitem[Fulkerson and Gross(1965)]{Delbert1965pjm-matrix-and-graph}
Delbert~Ray Fulkerson and Oliver~Alfred Gross.
\newblock Incidence matrices and interval graphs.
\newblock \emph{Pacific Journal of Mathematics}, 15:\penalty0 835--855, 1965.

\bibitem[Golumbic(2004)]{golumbic2004algorithmic}
Martin~Charles Golumbic.
\newblock \emph{Algorithmic graph theory and perfect graphs}.
\newblock Elsevier, 2004.

\bibitem[Graesdal et~al.(2024)Graesdal, Chia, Marcucci, Morozov, Amice, Parrilo, and Tedrake]{graesdal2024arxiv-tightconvexrelax-contactrich}
Bernhard~P Graesdal, Shao~YC Chia, Tobia Marcucci, Savva Morozov, Alexandre Amice, Pablo~A Parrilo, and Russ Tedrake.
\newblock Towards tight convex relaxations for contact-rich manipulation.
\newblock \emph{arXiv preprint arXiv:2402.10312}, 2024.

\bibitem[Hirai et~al.(1998)Hirai, Hirose, Haikawa, and Takenaka]{hirai1998icra-development-honda-humanoid}
Kazuo Hirai, Masato Hirose, Yuji Haikawa, and Toru Takenaka.
\newblock The development of honda humanoid robot.
\newblock In \emph{Proceedings. 1998 IEEE international conference on robotics and automation (Cat. No. 98CH36146)}, volume~2, pages 1321--1326. IEEE, 1998.

\bibitem[Huang et~al.(2024)Huang, Kang, Wang, and Yang]{huang2024arxiv-sparsehomogenization}
Lei Huang, Shucheng Kang, Jie Wang, and Heng Yang.
\newblock Sparse polynomial optimization with unbounded sets.
\newblock \emph{arXiv preprint arXiv:2401.15837}, 2024.

\bibitem[Kang et~al.(2024)Kang, Xu, Sarva, Liang, and Yang]{kang2024wafr-strom}
Shucheng Kang, Xiaoyang Xu, Jay Sarva, Ling Liang, and Heng Yang.
\newblock Fast and certifiable trajectory optimization.
\newblock \emph{arXiv preprint arXiv:2406.05846}, 2024.

\bibitem[Klep et~al.(2018)Klep, Povh, and Volcic]{klep2018siopt-minimizer-extraction-robust}
Igor Klep, Janez Povh, and Jurij Volcic.
\newblock Minimizer extraction in polynomial optimization is robust.
\newblock \emph{SIAM Journal on Optimization}, 28\penalty0 (4):\penalty0 3177--3207, 2018.

\bibitem[Koolen(2020)]{koolen2020arxiv-balance-control-humanoid-nonlinear-centroidal}
Frans~Anton Koolen.
\newblock \emph{Balance control and locomotion planning for humanoid robots using nonlinear centroidal models}.
\newblock PhD thesis, Massachusetts Institute of Technology, 2020.

\bibitem[Lasserre(2001)]{lasserre2001siopt-global}
Jean~B Lasserre.
\newblock Global optimization with polynomials and the problem of moments.
\newblock \emph{SIAM Journal on optimization}, 11\penalty0 (3):\penalty0 796--817, 2001.

\bibitem[Lasserre(2006{\natexlab{a}})]{Lasserre2006siam-convergent-sdp-relaxation}
Jean~B Lasserre.
\newblock Convergent sdp-relaxations in polynomial optimization with sparsity.
\newblock \emph{SIAM Journal on optimization}, 17\penalty0 (3):\penalty0 822--843, 2006{\natexlab{a}}.

\bibitem[Lasserre(2006{\natexlab{b}})]{lasserre2006msc-correlativesparse}
Jean~B Lasserre.
\newblock Convergent sdp-relaxations in polynomial optimization with sparsity.
\newblock \emph{SIAM Journal on optimization}, 17\penalty0 (3):\penalty0 822--843, 2006{\natexlab{b}}.

\bibitem[Le~Cleac'h et~al.(2024)Le~Cleac'h, Howell, Yang, Lee, Zhang, Bishop, Schwager, and Manchester]{le2024tro-fast-contact-implicit-mpc}
Simon Le~Cleac'h, Taylor~A Howell, Shuo Yang, Chi-Yen Lee, John Zhang, Arun Bishop, Mac Schwager, and Zachary Manchester.
\newblock Fast contact-implicit model predictive control.
\newblock \emph{IEEE Transactions on Robotics}, 2024.

\bibitem[Lee(2008)]{lee2008thesis-computationalgeometricmechanics}
Taeyoung Lee.
\newblock \emph{Computational geometric mechanics and control of rigid bodies}.
\newblock PhD thesis, University of Michigan, 2008.

\bibitem[Li et~al.(2025)Li, Han, Kang, Ma, and Yang]{li2025rss-crisp}
Yulin Li, Haoyu Han, Shucheng Kang, Jun Ma, and Heng Yang.
\newblock On the surprising robustness of sequential convex optimization for contact-implicit motion planning.
\newblock \emph{arXiv preprint arXiv:2502.01055}, 2025.

\bibitem[Lynch et~al.(1992)Lynch, Maekawa, and Tanie]{lynch1992iros-manipulation-active-sensing-pushing}
Kevin~M Lynch, Hitoshi Maekawa, and Kazuo Tanie.
\newblock Manipulation and active sensing by pushing using tactile feedback.
\newblock In \emph{IROS}, volume~1, pages 416--421, 1992.

\bibitem[Magron and Wang(2021)]{magron2021arxiv-julia-tssos}
Victor Magron and Jie Wang.
\newblock Tssos: a julia library to exploit sparsity for large-scale polynomial optimization.
\newblock \emph{arXiv preprint arXiv:2103.00915}, 2021.

\bibitem[Magron and Wang(2023)]{magron23book-sparse}
Victor Magron and Jie Wang.
\newblock \emph{Sparse polynomial optimization: theory and practice}.
\newblock World Scientific, 2023.

\bibitem[Manchester and Kuindersma(2020)]{manchester2020isrr-variational-contact-implicit}
Zachary Manchester and Scott Kuindersma.
\newblock Variational contact-implicit trajectory optimization.
\newblock In \emph{Robotics Research: The 18th International Symposium ISRR}, pages 985--1000. Springer, 2020.

\bibitem[Marcucci and Tedrake(2020)]{marcucci2020arxiv-warmstart-mixedinteger-mpc}
Tobia Marcucci and Russ Tedrake.
\newblock Warm start of mixed-integer programs for model predictive control of hybrid systems.
\newblock \emph{IEEE Transactions on Automatic Control}, 66\penalty0 (6):\penalty0 2433--2448, 2020.

\bibitem[Mason(1986)]{mason1986ijrr-mechanics-planning-pushing}
Matthew~T Mason.
\newblock Mechanics and planning of manipulator pushing operations.
\newblock \emph{The International Journal of Robotics Research}, 5\penalty0 (3):\penalty0 53--71, 1986.

\bibitem[Mastalli et~al.(2020)Mastalli, Budhiraja, Merkt, Saurel, Hammoud, Naveau, Carpentier, Righetti, Vijayakumar, and Mansard]{mastalli2020icra-crocoddyl}
Carlos Mastalli, Rohan Budhiraja, Wolfgang Merkt, Guilhem Saurel, Bilal Hammoud, Maximilien Naveau, Justin Carpentier, Ludovic Righetti, Sethu Vijayakumar, and Nicolas Mansard.
\newblock Crocoddyl: An efficient and versatile framework for multi-contact optimal control.
\newblock In \emph{2020 IEEE International Conference on Robotics and Automation (ICRA)}, pages 2536--2542. IEEE, 2020.

\bibitem[Mordatch et~al.(2012)Mordatch, Todorov, and Popovi{\'c}]{mordatch2012tog-discovery-complex-behaviors-contact-invariant}
Igor Mordatch, Emanuel Todorov, and Zoran Popovi{\'c}.
\newblock Discovery of complex behaviors through contact-invariant optimization.
\newblock \emph{ACM Transactions on Graphics (ToG)}, 31\penalty0 (4):\penalty0 1--8, 2012.

\bibitem[Morozov et~al.(2024)Morozov, Marcucci, Amice, Graesdal, Bosworth, Parrilo, and Tedrake]{morozov2024arxiv-multi-query-spp-gcs}
Savva Morozov, Tobia Marcucci, Alexandre Amice, Bernhard~Paus Graesdal, Rohan Bosworth, Pablo~A Parrilo, and Russ Tedrake.
\newblock Multi-query shortest-path problem in graphs of convex sets.
\newblock \emph{arXiv preprint arXiv:2409.19543}, 2024.

\bibitem[Olson(2011)]{olson2011icra-apriltag}
Edwin Olson.
\newblock Apriltag: A robust and flexible visual fiducial system.
\newblock In \emph{2011 IEEE international conference on robotics and automation}, pages 3400--3407. IEEE, 2011.

\bibitem[Pang et~al.(2023)Pang, Suh, Yang, and Tedrake]{pang2023tro-global-planning-contact-rich-quasi-dynamic-contact-models}
Tao Pang, HJ~Terry Suh, Lujie Yang, and Russ Tedrake.
\newblock Global planning for contact-rich manipulation via local smoothing of quasi-dynamic contact models.
\newblock \emph{IEEE Transactions on robotics}, 2023.

\bibitem[Posa et~al.(2014)Posa, Cantu, and Tedrake]{posa2014ijrr-traopt-directmethod-contact}
Michael Posa, Cecilia Cantu, and Russ Tedrake.
\newblock A direct method for trajectory optimization of rigid bodies through contact.
\newblock \emph{The International Journal of Robotics Research}, 33\penalty0 (1):\penalty0 69--81, 2014.

\bibitem[Rose et~al.(1976)Rose, Tarjan, and Lueker]{Rose1976siam-vertex-elimination}
Donald~J Rose, R~Endre Tarjan, and George~S Lueker.
\newblock Algorithmic aspects of vertex elimination on graphs.
\newblock \emph{SIAM Journal on computing}, 5\penalty0 (2):\penalty0 266--283, 1976.

\bibitem[Tassa et~al.(2012)Tassa, Erez, and Todorov]{tassa2012iros-synthesis-stabilization-online-traopt}
Yuval Tassa, Tom Erez, and Emanuel Todorov.
\newblock Synthesis and stabilization of complex behaviors through online trajectory optimization.
\newblock In \emph{2012 IEEE/RSJ International Conference on Intelligent Robots and Systems}, pages 4906--4913. IEEE, 2012.

\bibitem[Tassa et~al.(2014)Tassa, Mansard, and Todorov]{tassa2014icra-control-limitted-ddp}
Yuval Tassa, Nicolas Mansard, and Emo Todorov.
\newblock Control-limited differential dynamic programming.
\newblock In \emph{2014 IEEE International Conference on Robotics and Automation (ICRA)}, pages 1168--1175. IEEE, 2014.

\bibitem[Teng et~al.(2023)Teng, Jasour, Vasudevan, and Ghaffari]{teng2023arxiv-geometricmotionplanning-liegroup}
Sangli Teng, Ashkan Jasour, Ram Vasudevan, and Maani Ghaffari.
\newblock Convex geometric motion planning on lie groups via moment relaxation.
\newblock In \emph{Robotics: Science and Systems}, 2023.

\bibitem[Teng et~al.(2024)Teng, Jasour, Vasudevan, and Ghaffari]{teng2024ijrr-convex-geometric-motion-planning}
Sangli Teng, Ashkan Jasour, Ram Vasudevan, and Maani Ghaffari.
\newblock Convex geometric motion planning of multi-body systems on lie groups via variational integrators and sparse moment relaxation.
\newblock \emph{The International Journal of Robotics Research}, page 02783649241296160, 2024.

\bibitem[Waki et~al.(2006)Waki, Kim, Kojima, and Muramatsu]{Waki2006siam-sos-semidefinite-relaxation}
Hayato Waki, Sunyoung Kim, Masakazu Kojima, and Masakazu Muramatsu.
\newblock Sums of squares and semidefinite program relaxations for polynomial optimization problems with structured sparsity.
\newblock \emph{SIAM Journal on Optimization}, 17\penalty0 (1):\penalty0 218--242, 2006.

\bibitem[Wang(2023)]{wang2023book-introduction-pop}
Jie Wang.
\newblock An introduction to polynomial optimization.
\newblock 2023.

\bibitem[Wang et~al.(2021)Wang, Magron, and Lasserre]{wang2021siam-tssos}
Jie Wang, Victor Magron, and Jean-Bernard Lasserre.
\newblock Tssos: A moment-sos hierarchy that exploits term sparsity.
\newblock \emph{SIAM Journal on optimization}, 31\penalty0 (1):\penalty0 30--58, 2021.

\bibitem[Wang et~al.(2022)Wang, Magron, Lasserre, and Mai]{wang2022tms-cs-tssos}
Jie Wang, Victor Magron, Jean~B Lasserre, and Ngoc Hoang~Anh Mai.
\newblock Cs-tssos: Correlative and term sparsity for large-scale polynomial optimization.
\newblock \emph{ACM Transactions on Mathematical Software}, 48\penalty0 (4):\penalty0 1--26, 2022.

\bibitem[Wu et~al.(2020)Wu, Sadraddini, and Tedrake]{wu2020icra-r3t-nonlinear-hybrid}
Albert Wu, Sadra Sadraddini, and Russ Tedrake.
\newblock R3t: Rapidly-exploring random reachable set tree for optimal kinodynamic planning of nonlinear hybrid systems.
\newblock In \emph{2020 IEEE International Conference on Robotics and Automation (ICRA)}, pages 4245--4251. IEEE, 2020.

\bibitem[Yang(2024)]{Yang2024book-sdp}
Heng Yang.
\newblock Semidefinite optimization and relaxation.
\newblock Lecture notes: \url{https://hankyang.seas.harvard.edu/Semidefinite/}, 2024.

\bibitem[Yang and Carlone(2022)]{yang2022pami-outlierrobust-geometricperception}
Heng Yang and Luca Carlone.
\newblock Certifiably optimal outlier-robust geometric perception: Semidefinite relaxations and scalable global optimization.
\newblock \emph{IEEE transactions on pattern analysis and machine intelligence}, 45\penalty0 (3):\penalty0 2816--2834, 2022.

\bibitem[Yang et~al.()Yang, Marcucci, Parrilo, and Tedrake]{yang2024arxiv-sdp-linear-piecewise-affine-optimal-control}
Lujie Yang, Tobia Marcucci, Pablo~A Parrilo, and Russ Tedrake.
\newblock A new semidefinite relaxation for linear and piecewise-affine optimal control with time scaling.

\bibitem[Yang and Posa(2024)]{yang2024rss-dynamic-on-plam-control-sliding}
William Yang and Michael Posa.
\newblock Dynamic on-palm manipulation via controlled sliding.
\newblock \emph{arXiv preprint arXiv:2405.08731}, 2024.

\bibitem[Yannakakis(1981)]{Yannakakis1981siam-minimum-fill-in}
Mihalis Yannakakis.
\newblock Computing the minimum fill-in is np-complete.
\newblock \emph{SIAM Journal on Algebraic Discrete Methods}, 2\penalty0 (1):\penalty0 77--79, 1981.

\bibitem[Yunt(2006)]{yunt2006-opttraj-planning-structure-variant}
K.~Yunt.
\newblock Optimal trajectory planning for structure-variant mechanical systems.
\newblock In \emph{International Workshop on Variable Structure Systems, 2006. VSS'06.}, pages 298--303, 2006.
\newblock \doi{10.1109/VSS.2006.1644534}.

\bibitem[Yunt and Glocker(2007)]{yunt2007isdc-combined-continuation-penalty}
Kerim Yunt and Christoph Glocker.
\newblock A combined continuation and penalty method for the determination of optimal hybrid mechanical trajectories.
\newblock In \emph{Iutam Symposium on Dynamics and Control of Nonlinear Systems with Uncertainty: Proceedings of the IUTAM Symposium held in Nanjing, China, September 18-22, 2006}, pages 187--196. Springer, 2007.

\end{thebibliography}

\clearpage
\onecolumn
\appendix

\renewcommand{\theequation}{A\arabic{equation}} % Prefix "A" to equation numbers
\setcounter{equation}{0} % Reset equation counter
\renewcommand{\thefigure}{A\arabic{figure}} % Prefix "A" to figures
\setcounter{figure}{0} % Reset figure counter
\renewcommand{\thetheorem}{A\arabic{theorem}} % Prefix "A"
\setcounter{theorem}{0} % Reset theorem counter

%!TEX root = ../main.tex
\subsection{MD and MF Chordal Extension}
\label{app:sec:mdmf}

We present the MD chordal extension algorithm, which selects vertices based on minimum degree for elimination ordering.

\begin{algorithm}
    \small 
    \SetAlgoLined
    \caption{MD Chordal Extension ~\label{alg:gs:md}}
    \KwIn{Graph $G(V,E)$ with $n$ vertices}
    \KwOut{Chordal graph $G'$, elimination order $\pi$}
    $G' \leftarrow G$, $H \leftarrow G$, $R \leftarrow V$\;
    \For{$i = 1$ to $n$}{
        Find vertex $v \in R$ with minimum degree in $H$\;
        $\pi(v) \leftarrow i$\;
        \texttt{// Unprocessed neighbors}\;
        $N \leftarrow \{u \in R : (v,u) \in E(H)\}$\;
        \For{each pair $(u,w) \in N \times N, u \neq w$}{
            \If{$(u,w) \notin E'$}{
                Add edge $(u,w)$ to $G'$ and $H$\;
                Remove edges $(v, u)$ in H, $\forall u \in N$\;
            }
        }
        $R \leftarrow R \setminus \{v\}$\;
    }
\end{algorithm}

We also present the MF chordal extension algorithm, which selects vertices causing minimum fill-in edges during the elimination process.

\begin{algorithm}
    \small 
    \SetAlgoLined
    \caption{MF Chordal Extension ~\label{alg:gs:mf}}
    \KwIn{Graph $G(V,E)$ with $n$ vertices}
    \KwOut{Chordal graph $G'$, elimination order $\pi$}
    $G' \leftarrow G$, $H \leftarrow G$, $R \leftarrow V$\;
    \For{$i = 1$ to $n$}{
        For each $v \in R$, compute fill-in cost:
        $\text{fill-in}(v) = |\{(u,w) : u,w \in N_H(v), u \neq w, (u,w) \notin E(H)\}|$\;
        Find vertex $v \in R$ with minimum fill-in$(v)$ in $H$\;
        $\pi(v) \leftarrow i$\;
        \texttt{// Unprocessed neighbors}\;
        $N \leftarrow \{u \in R : (v,u) \in E(H)\}$\;
        \For{each pair $(u,w) \in N \times N, u \neq w$}{
            \If{$(u,w) \notin E'$}{
                Add edge $(u,w)$ to $G'$ and $H$\;
                Remove edges $(v, u)$ in H, $\forall u \in N$\;
            }
        }
        $R \leftarrow R \setminus \{v\}$\;
    }
\end{algorithm}
%!TEX root = ../main.tex

\subsection{Moment-SOS Hierarchy with CS-TS}
\label{app:sec:cs-ts-relax}

Given a graph $G(V, E)$, define:
\begin{align}
\mbS_G = \cbrace{Q\in \mbS^{|V|}\mymid Q_{\beta,\gamma} = Q_{\gamma,\beta} = 0, \forall \beta\ne\gamma, (\beta, \gamma)\notin E}
\end{align}
where the rows and columns of $Q \in \mathbf{S}_G$ are indexed by $V$. Let $\Pi_G$ be the projection form $\mbS^{|V|}$ to the subspace $\mbS_G$. Specifically, forall $Q\in \mbS^{|V|}$:
\begin{align}
	\Pi_G(Q) = \begin{cases}
		Q_{\beta,\gamma}, & \beta = \gamma \text{ or } (\beta, \gamma)\in E\\
		0, & \text{otherwise}
	\end{cases}
\end{align}
If we further define $\Pi_G(\mathbf{S}^{|V|}_+)$ as $\cbrace{\Pi_G(Q) \mymid Q \in \mathbf{S}^{|V|}_+}$, the image of PSD cone with under projection $\Pi_G(\cdot)$, then the Moment-SOS Hierarchy with combined CS and TS can be concretely written as:
\begin{eqnarray}
    \label{eq:gs-cs-ts}
    \min & L_y(f) \\
    \text{s.t.} & L_y\left( M_d(g_j, I_l) \right) \circ B_{d,l,j}^g \in \Pi_{G'_{d,l,j}}(S^{|V_{d,l,j}|}_{+}), \nonumber \\
    & \forall j\in \cbrace{0} \cup \calG_l, l\in [p] \\
    & L_y\left( H_d(h_j, I_l) \right) \circ B_{d,l,j}^h = 0, \forall j\in \calH_l, l\in [p] \\
    & y_{\mathbf{0}} = 1 
\end{eqnarray} 

The above procedure outlines the CS-TS Moment-SOS Hierarchy with a sparse order of $k = 1$. As shown in~\cite{wang2022tms-cs-tssos}, one can further iteratively apply support extension and chordal extension within term sparsity. This process generates new sets $B_{d,l,j}^g$ and $B_{d,l,j}^h$, leading to a two-level hierarchy:  
\begin{enumerate}
    \item The outer level is governed by CS's relaxation order $d$.  
    \item The inner level is controlled by TS's sparse order $k$, which corresponds to the number of iterations used to generate new $B_{d,l,j}^g$ and $B_{d,l,j}^h$.  
\end{enumerate}
Define the optimal value of~\eqref{eq:gs-cs-ts} as $\rho_d^k$. The sequence $\{\rho_d^k\}_{k \geq 1}$ is monotonically non-decreasing and satisfies $\rho_d^k \leq \rho_d$ for all $k$. What's more:

\begin{theorem}[Theorem 4.26 in~\cite{wang2023book-introduction-pop}]
    If we use maximal chordal extension (\ie block closure) for term sparsity, $\rho_d^k \rightarrow \rho_d$ as $k \rightarrow \infty$.
\end{theorem}
%!TEX root = ../main.tex
\subsection{Polynomial Dynamics of Robotics Systems}
\label{app:sec:polynomial-dynamics}

\subsubsection{Push Bot}
\label{app:pd:push-bot}
Push bot is essentially cart-pole with soft wall. The configuration is shown in Figure~\ref{fig:exp:illustration} (a). $a$ is cart's position, $\theta$ is pole's angle, $k_1$ and $k_2$ is soft wall's elastic modulus, $\lamone$ and $\lamtwo$ is two contact forces between two walls and pole's tip. The goal is to stabilize the cart-pole to $(a,\theta) = (0, \pi)$.
From Newtonian mechanics:
\begin{align}
    \label{eq:app:pd:push-bot-newton}
    & (\mcar + \mpole) \frac{d^2}{dt^2}\pos + \mpole \len \frac{d^2}{dt^2}(\sin\theta) - (u + \lamone - \lamtwo) = 0 \\
    & \len \frac{d^2}{dt^2}\theta + (\frac{d^2}{dt^2}\pos + \lamtwo - \lamone) \cos\theta + g \sin\theta = 0 \\
    & 0 \le \lamone \perp \frac{\lamone}{k_1} + d_1 + (\pos + \len \sin\theta) \ge 0 \\
    & 0 \le \lamtwo \perp \frac{\lamtwo}{k_2} + d_2 - (\pos + \len \sin\theta) \ge 0 
\end{align}
Use the same techniques introduced in~\cite{teng2023arxiv-geometricmotionplanning-liegroup}, we discretize~\eqref{eq:app:pd:push-bot-newton} on the lie group~\cite{lee2008thesis-computationalgeometricmechanics} to yield polynomial dynamics:
\begin{align}
    \label{eq:app:pd:push-bot-liegroup}
    & (\mcar + \mpole) \cdot \frac{\pos[k+1] - 2\pos[k] + \pos[k-1]}{\dt^2} \nonumber \\ 
    & + (\mpole \len) \cdot \frac{\rs[k+1] - 2 \rs[k] + \rs[k-1]}{\dt^2} - (u_k + \lamone[k] - \lamtwo[k]) = 0 \\
    & \len \cdot \frac{\fs[k] - \fs[k-1]}{\dt^2} \nonumber \\
    & + \left( 
        \frac{\pos[k+1] - 2\pos[k] + \pos[k-1]}{\dt^2} + (\lamtwo[k] - \lamone[k])
     \right) \cdot \rc[k] + g \cdot \rs[k] = 0 \\
    & 0 \le \lamone[k] \perp \left( \frac{\lamone[k]}{k_1} + d_1 + \pos[k] + \len \rs[k] \right) \ge 0 \\
    & 0 \le \lamtwo[k] \perp \left( \frac{\lamtwo[k]}{k_2} + d_2 - \pos[k] - \len \rs[k] \right) \ge 0 \\
    & \rc[k] = \rc[k-1] \fc[k-1] - \rs[k-1] \fs[k-1] \label{eq:app:pd:2d-liegroup-start} \\
    & \rs[k] = \rs[k-1] \fc[k-1] + \rc[k-1] \fs[k-1] \\
    & \rc[k]^2 + \rs[k]^2 = 1 \\
    & \fc[k]^2 + \fs[k]^2 = 1 \label{eq:app:pd:2d-liegroup-end}
\end{align}
The loss function is designed as:
\begin{align}
    \label{eq:app:pd:push-bot-loss}
    & \loss = \sum_{k=0}^{N-1} c_a \cdot a_k^2 + c_{a,f} \cdot a_N^2 \nonumber \\
        & + \sum_{k=0}^{N-1} c_\theta \cdot \left\{ 
            (\rc[k] + 1)^2 + \rs[k]^2 
         \right\} + c_{\theta,f} \cdot \left\{ 
            (\rc[N] + 1)^2 + \rs[N]^2 
          \right\} \nonumber \\
        & + \sum_{k=0}^{N-1} c_{\dot{\theta}} \cdot \left\{ 
            (\fc[k] - 1)^2 + \fs[k]^2
          \right\} + c_{\dot{\theta},f} \cdot \left\{ 
            (\fc[N] - 1)^2 + \fs[N]^2
           \right\}
\end{align}

\subsubsection{Push Box}
\label{app:pd:push-box}
Consider a simple pusher-slider system illustrated in Figure~\ref{fig:exp:illustration} (b). Our goal is to push the box from one configuration ($(\sx, \sy, \theta)$) to another.
From~\cite{graesdal2024arxiv-tightconvexrelax-contactrich}, given (1) the pusher's position $(\px, \py)$ and the contact force $(\Fx, \Fy)$ in the slider frame; (2) the slider's position $(\sx, \sy)$ and angle $\theta$ in the world frame, the quasi-static dynamics of the slider can be written as:
\begin{align}
    \label{eq:app:pd:push-box-quasi-static}
    & \frac{d}{dt} \sx = \frac{1}{\mu_1 mg} \cdot (\cos\theta \Fx - \sin\theta \Fy) \\
    & \frac{d}{dt} \sy = \frac{1}{\mu_1 mg} \cdot (\sin\theta \Fx + \cos\theta \Fy) \\
    & \frac{d}{dt} \theta = \frac{1}{c r \cdot \mu_1 mg} \cdot (-\py \Fx + \px \Fy) 
\end{align}
where $\mu_1$ is the friction coefficient between the slider and table. $c \in (0, 1)$ is the integration constant that depends on the slider geometry. $r$ is a characteristic distance, typically chosen as the max distance between a contact point and origin of slider frame~\cite{mason1986ijrr-mechanics-planning-pushing}. Use the dimensionless trick:
\begin{align}
    \Fx \leftarrow \frac{1}{\mu_1 mg} \cdot \Fx, \ \Fy \leftarrow \frac{1}{\mu_1 mg} \cdot \Fy
\end{align}
Discretize over the lie group:
\begin{align}
    & \sx[k] = \sx[k-1] + \dt \cdot (\rc[k-1] \Fx[k-1] - \rs[k-1] \Fy[k-1]) \\
    & \sy[k] = \sy[k-1] + \dt \cdot (\rs[k-1] \Fx[k-1] + \rc[k-1] \Fy[k-1]) \\
    & \fs[k-1] = \dt \cdot \frac{1}{cr} \cdot (-\py[k-1] \Fx[k-1] + \px[k-1] \Fy[k-1]) 
\end{align}
Here, the lie-group constraints~\eqref{eq:app:pd:2d-liegroup-start} -~\eqref{eq:app:pd:2d-liegroup-end} are omitted for simplicity. 
Since when pusher has no contact with the slider, slider remains still and pusher's planning task is trivial, we only fucus on the time steps when pusher and slider have contact. As illustrated in Figure~\ref{fig:exp:illustration} (b), we assign for modes $\lam{i}$'s ($ i = 1, 2, 3, 4$ for box's four sides):
\begin{align}
    & \lam{i} (1 - \lam{i}) = 0, \ i = 1, 2, 3, 4 \\
    & \sum_{i=1}^4 \lam{i}^2 = 1
\end{align}
In each mode, the relationship between $\Fx, \Fy, \px, \py$ is different. For example, in mode 1:
\begin{align}
    & \lam{1} \cdot (a^2 - \px^2) \ge 0 \\
    & \lam{1} \cdot (\py - b) = 0 \\
    & \lam{1} \cdot (-\Fy) \ge 0 
\end{align}
where for modelling simplicity, we assume the pushing direction will always be normal to the contact surface. Similar contact constraints can be assigned to mode 2 - 4. Simplify them:
\begin{align}
    & (\lam{1} + \lam{3}) \cdot (a^2 - \px^2) + (\lam{2} + \lam{4}) \cdot (b^2 - \py^2) \ge 0 \\
    & \lam{1} \cdot (\py - b) + \lam{2} \cdot (\px - a) + \lam{3} \cdot (\py + b) + \lam{4} \cdot (\px + a) = 0 \\
    & (-\lam{1} + \lam{3}) \cdot \Fy + (-\lam{2} + \lam{4}) \cdot \Fx \ge 0 \\
    & (\lam{1} + \lam{3}) \cdot \Fx + (\lam{2} + \lam{4}) \cdot \Fy = 0 
\end{align}
The loss function is in the same spirit as~\eqref{eq:app:pd:push-bot-loss}. We omit it here.

\subsubsection{Push T-block}
\label{app:pd:push-T-block}

Now we consider a more complicated pushing task: push a T-block, as illustrated in Figure~\ref{fig:exp:illustration} (c). Unlike 4 modes in the box setting, now we have 8 modes to assign. From~\cite{lynch1992iros-manipulation-active-sensing-pushing}, when $\mu_1$ is uniformly distributed between the slider and the table, the friction center coincides with the projection of the center of mass (CM) to the table. Thus, we set the origin of the Slider frame to T-block's CM for convenience. $d_c$ from Figure~\ref{fig:exp:illustration} (c) can be derived as:
\begin{align}
    d_c = \frac{
        3 \times 1.5 + 4 \times 3.5
    }{3 + 4} = \frac{37}{14}
\end{align}
There are eight key points in the T-block:
\begin{align}
    & x_1 = -2l, \ x_2 = -0.5l, \ x_3 = 0.5l, \ x_4 = 2l \\
    & y_1 = -d_c l, \ y_2 = (3 - d_c) l, \ y_3 = (4 - d_c) l 
\end{align}
Connect each mode with geometric and dynamical constraints:
\begin{align}
    & \lam{1} \Longrightarrow \py - y_3 = 0, \px - x_1 \ge 0, x_4 - \px \ge 0, -\Fy \ge 0, \Fx = 0 \\
    & \lam{2} \Longrightarrow \px - x_4 = 0, \py - y_2 \ge 0, y_3 - \py \ge 0, -\Fx \ge 0, \Fy = 0 \\
    & \lam{3} \Longrightarrow \py - y_2 = 0, \px - x_3 \ge 0, x_4 - \px \ge 0, \Fy \ge 0, \Fx = 0 \\
    & \lam{4} \Longrightarrow \px - x_3 = 0, \py - y_1 \ge 0, y_2 - \py \ge 0, -\Fx \ge 0, \Fy = 0 \\
    & \lam{5} \Longrightarrow \py - y_1 = 0, \px - x_2 \ge 0, x_3 - \px \ge 0, \Fy \ge 0, \Fx = 0 \\
    & \lam{6} \Longrightarrow \px - x_2 = 0, \py - y_1 \ge 0, y_2 - \py \ge 0, \Fx \ge 0, \Fy = 0 \\
    & \lam{7} \Longrightarrow \py - y_2 = 0, \px - x_1 \ge 0, x_2 - \px \ge 0, \Fy \ge 0, \Fx = 0 \\
    & \lam{8} \Longrightarrow \px - x_1 = 0, \py - y_2 \ge 0, y_3 - \py \ge 0, \Fx \ge 0, \Fy = 0
\end{align}
Other things are the same as the Push Box case.

\subsubsection{Push Box with a Tunnel}
\label{app:pd:push-box-tunnel}

Everything is the same as Push Box setting, except that the box needs to avoid two circle obstacles this time. To model the collision avoidance constraints, we approximate the box as a union of two circles, as shown in Figure~\ref{fig:exp:illustration} (d). For each obstacle-slider circle pair, the non-collision constraint is:
\begin{align}
    (x_o - x_s)^2 + (y_o - y_s)^2 \ge (r_o + r_s)^2
\end{align}
where $(x_o, y_o, r_o)$ (resp. $(x_s, y_s, r_s)$ ) represents center and radius of obstacle's (resp. slider's) center.

\subsubsection{Planar Hand}
\label{app:pd:planar-hand}

The geometric and mechanical information of the Planar Hand system is illustrated in Figure~\ref{fig:exp:illustration} (e). The goal is to rotate the circle disk $360^\circ$ with planar hand's two finger tips, while minimize the translation of the disk's center of mass. 

\textbf{Kinematics of the fingers.}
For two fingers, we use position control. For example, for the right finger:
\begin{align}
    & x_r = L_d \cdot \cos\theta_{rd} + L_u \cdot \cos\theta_{ru} + \frac{H}{2} \\
    & y_r = L_d \cdot \sin\theta_{rd} + L_u \cdot \sin\theta_{ru} \\
    & v_{x,r} = -L_d \cdot \sin\theta_{rd} \cdot \dot{\theta}_{rd} - L_u \cdot \sin\theta_{ru} \cdot \dot{\theta}_{ru} \\
    & v_{y,r} = L_d \cdot \cos\theta_{rd} \cdot \dot{\theta}_{rd} + L_u \cdot \cos\theta_{ru} \cdot \dot{\theta}_{ru}
\end{align}
where "r" and "l" represent "right" and "left" finger, while "u" and "d" represent "upper" and "down" link. Since 
\begin{align}
    \fs = \sin(\dot{\theta} \cdot \dt) \Longrightarrow \dot{\theta} \approx \frac{\fs}{\dt}
\end{align}
Then,
\begin{align}
    & \xr[k] = L_d \cdot \rc[rd,k] + L_u \cdot \rc[ru,k] + \frac{H}{2} \\
    & \yr[k] = L_d \cdot \rs[rd,k] + L_u \cdot \rs[ru,k] \\
    & \vxr[k] = -\frac{L_d}{\dt} \cdot \rs[rd,k] \cdot \fs[rd,k] - \frac{L_u}{\dt} \cdot \rs[ru,k] \cdot \fs[ru,k] \\
    & \vyr[k] = \frac{L_d}{\dt} \cdot \rc[rd,k] \cdot \fs[rd,k] + \frac{L_u}{\dt} \cdot \rc[ru,k] \cdot \fs[ru,k] \\
    & \rc[rd,k+1] = \rc[rd,k] \cdot \fc[rd,k] - \rs[rd,k] \cdot \fs[rd,k] \\
    & \rs[rd,k+1] = \rc[rd,k] \cdot \fs[rd,k] + \rs[rd,k] \cdot \fc[rd,k] \\
    & \rc[ru,k+1] = \rc[ru,k] \cdot \fc[ru,k] - \rs[ru,k] \cdot \fs[ru,k] \\
    & \rs[ru,k+1] = \rc[ru,k] \cdot \fs[ru,k] + \rs[ru,k] \cdot \fc[ru,k]
\end{align}

\textbf{Self collision avoidance.}
For each finger, there are two types of self collisions: (1) the first circle with the ground; (2) the second and the third circle. For the first type:
\begin{align}
    & \theta_{ld} \ge \arcsin(\frac{r}{l+r}), \ \pi - \theta_{ld} \ge \arcsin(\frac{r}{l+r}) \\
    & \theta_{rd} \ge \arcsin(\frac{r}{l+r}), \ \pi - \theta_{rd} \ge \arcsin(\frac{r}{l+r})
\end{align}
For the second type:
\begin{align}
    & \pi - \theta_{ld} + \theta_{lu} \ge 2 \cdot \arcsin(\frac{r}{l+r}) \\
    & 2 \pi - (\pi - \theta_{rd} + \theta_{ru}) \ge 2 \cdot \arcsin(\frac{r}{l+r}) 
\end{align}
Also, like a human finger, we assume the upper link won't "turn outward":
\begin{align}
    & \theta_{ld} - \theta_{lu} \ge 0 \\
    & \theta_{ru} - \theta_{rd} \ge 0 
\end{align}

Now denote $\theta_0$ as $\arcsin(\frac{r}{l+r})$. Writing the constraints as polynomials:
\begin{align}
    & \rs[ld,k] \ge \sin\theta_0 \\
    & \rs[rd,k] \ge \sin\theta_0
\end{align}
and 
\begin{align}
    & \sin(\theta_{ld,k} - \theta_{lu,k}) 
    = \rs[ld,k] \cdot \rc[lu,k] - \rc[ld,k] \cdot \rs[lu,k] \ge 0 \\
    & \cos(\theta_{ld,k} - \theta_{lu,k}) 
    = \rc[ld,k] \cdot \rc[lu,k] + \rs[ld,k] \cdot \rs[lu,k] \ge -\cos(2\theta_0) \\
    & \sin(\theta_{ru,k} - \theta_{rd,k}) 
    = \rs[ru,k] \cdot \rc[rd,k] - \rc[ru,k] \cdot \rs[rd,k] \ge 0 \\
    & \cos(\theta_{ru,k} - \theta_{rd,k}) 
    = \rc[ru,k] \cdot \rc[rd,k] + \rs[ru,k] \cdot \rs[rd,k] \ge -\cos(2\theta_0)
\end{align}

\textbf{Contact model.}
Now we deal with the contact between the fingers and the disk. Without loss of generality, we consider the right finger. Denote $d_r$ as:
\begin{align}
    & d_r^2 = (x_r - x)^2 + (y_r - y)^2 \\
    & d_r \ge R+r
\end{align}
where $(x, y)$ is the position of the disk's center. When contact happens, $d_r = R+r$. In this case, denote $(v_x, v_y, w)$ as the translational and angular velocity of the disk, $(\lam{t}[r], \lam{n}[r])$ as the tangential and normal force exerted on the disk by the tip of the finger, and $\vrelr$ as the relative tangential velocity of finger's tip compared to the disk. The physical quantities are illustrated in Figure~\ref{fig:exp:illustration} (e)'s upper right position. Denote the angle $\eta_r$ as:
\begin{align}
    & \cos\eta_r = \frac{x_r - x}{d_r} \\
    & \sin\eta_r = \frac{y_r - y}{d_r} 
\end{align}
Then,
\begin{align}
    \vrelr \approx & -\vxr \cdot \sin\eta_r + \vyr \cdot \cos\eta_r \nonumber \\
    & - (-v_x \cdot \sin\eta_r + v_y \cdot \cos\eta_r + \omega R)
\end{align}
Here we do one approximation for finger tip's tangential velocity, by ignoring the tip's size. One thing we should notice is, $\lam{n}$ will always point inside the disk:
\begin{align}
    -\lam{n,r} \ge 0 
\end{align}
By the Coulomb's law:
\begin{align}
    \lam{t,r} \begin{cases}
        & = \mu \cdot (-\lam{n,r}), \ \vrelr > 0 \\
        & = -\mu \cdot (-\lam{n,r}), \ \vrelr < 0 \\
        & \in [-\mu \cdot (-\lam{n,r}), \mu \cdot (-\lam{n,r})], \ \vrelr = 0
    \end{cases}
\end{align}
Unlike~\cite{posa2014ijrr-traopt-directmethod-contact} who introduced two auxiliary variables to app:pdress the above model as nine quadratic polynomials, we advocate for a more concise representation:
\begin{align}
    & \mu^2 \cdot \lam{n,r}^2 - \lam{t,r}^2 \ge 0 \\
    & \vrelr \cdot (\mu^2 \cdot \lam{n,r}^2 - \lam{t,r}^2) = 0 \\
    & \vrelr \cdot \lam{t,r} \ge 0 
\end{align}
Combine everything together:
\begin{align}
    & \dr[k]^2 = (\xr[k] - \x[k])^2 + (\yr[k] - \y[k])^2 \\
    & \dr[k] \ge R + r \\
    & \vrelr[k] = -(\vxr[k] - \vx[k]) \cdot \frac{\yr[k] - \y[k]}{R+r} \nonumber \\
    & + (\vyr[k] - \vy[k]) \cdot \frac{\xr[k] - \x[k]}{R+r} 
    - R \cdot \frac{\fs[k]}{\dt} \\
    & (\dr[k] - R - r) \cdot \lam{n,r}[k] = 0 \\
    & -\lam{n,r}[k] \ge 0 \\
    & \mu^2 \cdot \lam{n,r}[k]^2 - \lam{t,r}[k]^2 \ge 0 \\
    & \vrelr[k] \cdot (\mu^2 \cdot \lam{n,r}[k]^2 - \lam{t,r}[k]^2) = 0 \\
    & \vrelr[k] \cdot \lam{t,r}[k] \ge 0 
\end{align}

\textbf{Dynamics of disk.}
Consider the quasi-static dynamics similar to~\eqref{eq:app:pd:push-box-quasi-static}:
\begin{align}
    & \frac{d}{dt} \x = \lam{n,r} \cdot \cos\eta_r - \lam{t,r} \cdot \sin\eta_r + \lam{n,l} \cdot \cos\eta_l - \lam{t,r} \cdot \sin\eta_l \\
    & \frac{d}{dt} \y = \lam{n,r} \cdot \sin\eta_r + \lam{t,r} \cdot \cos\eta_r + \lam{n,l} \cdot \sin\eta_l + \lam{t,r} \cdot \cos\eta_l \\
    & \frac{d}{dt} \alpha = \frac{1}{c \cdot R} \cdot (\lam{t,r} + \lam{t,l})
\end{align}
Write them as polynomials:
\begin{align}
    & \frac{1}{\dt} (\x[k+1] - \x[k]) = \lam{n,r}[k] \cdot \frac{\xr[k] - \x[k]}{R+r} - \lam{t,r}[k] \cdot \frac{\yr[k] - \y[k]}{R+r} 
    + \lam{n,l}[k] \cdot \frac{\xl[k] - \x[k]}{R+r} - \lam{t,l}[k] \cdot \frac{\yl[k] - \y[k]}{R+r} \\
    & \frac{1}{\dt} (\y[k+1] - \y[k]) = \lam{n,r}[k] \cdot \frac{\yr[k] - \y[k]}{R+r} + \lam{t,r}[k] \cdot \frac{\xr[k] - \x[k]}{R+r} 
    + \lam{n,l}[k] \cdot \frac{\yl[k] - \y[k]}{R+r} + \lam{t,l}[k] \cdot \frac{\xl[k] - \x[k]}{R+r} \\
    & \fs[k] = \frac{\dt}{c \cdot R} \cdot (\lam{t,r}[k] + \lam{t,l}[k]) \\
    & \rc[k+1] = \rc[k] \cdot \fc[k] - \rs[k] \cdot \fs[k] \\
    & \rs[k+1] = \rc[k] \cdot \fs[k] + \rs[k] \cdot \fc[k] \\
    & \fs[k]^2 + \fc[k]^2 = 1 
\end{align}

\textbf{Collision avoidance.}
Consider the right finger. Since the finger tip has already been considered in the contact model, we only need to consider collision avoidance between the object and the remaining three circles attached to the finger. The three circles' positions are:
\begin{align}
    & \left( 
        (l+r) \cos\theta_{rd} + \frac{H}{2}, (l+r) \sin\theta_{rd}
     \right) \\
    & \left( 
        (2l + 3r) \cos\theta_{rd} + \frac{H}{2}, (2l + 3r) \sin\theta_{rd}
    \right) \\
    & \left( 
        L_d \cos\theta_{rd} + (l+r) \cos\theta_{ru} + \frac{H}{2}, L_d \sin\theta_{rd} + (l+r) \sin\theta_{ru}
     \right)
\end{align}
Therefore, the constraints are:
\begin{align}
    & \left( (l+r) \cdot \rc[rd,k] - \x[k]  + \frac{H}{2} \right)^2 
    + \left( (l+r) \cdot \rs[rd,k] - \y[k] \right)^2 \ge (R+r)^2 \\
    & \left( (2l+3r) \cdot \rc[rd,k] - \x[k]  + \frac{H}{2} \right)^2 
    + \left( (2l+3r) \cdot \rs[rd,k] - \y[k] \right)^2 \ge (R+r)^2 \\
    & \left( L_d \cdot \rc[rd,k] + (l+r) \cdot \rc[ru,k] - \x[k]
    + \frac{H}{2} \right)^2 
    + \left( L_d \cdot \rs[rd,k] + (l+r) \cdot \rs[ru,k] - \y[k] \right)^2 \ge (R+r)^2
\end{align}

%!TEX root = ../main.tex
\subsection{Self-Defined Variable Cliques for Planar hand}
\label{app:sec:self-cliques}

The self-defined variable cliques for PLanar Hand are defined as:
\begin{subequations}
    \label{eq:exp:planar-hand-cliques}
    \begin{align}
        & \left\{ x_{k}, y_{k}, r_{c, k}, r_{s, k}, r_{c, ld, k}, r_{s, ld, k}, r_{c, lu, k}, r_{s, lu, k}, r_{c, rd, k}, r_{s, rd, k}, r_{c, ru, k}, r_{s, ru, k} \right\} \\ 
        & \left\{ x_{k}, y_{k}, r_{c, ld, k}, r_{s, ld, k}, r_{c, lu, k}, r_{s, lu, k}, x_{l, k}, y_{l, k}, d_{l, k} \right\} \\ 
        & \left\{ r_{c, ld, k}, r_{s, ld, k}, r_{c, lu, k}, r_{s, lu, k}, x_{l, k}, y_{l, k} \right\} \\ 
        & \left\{ r_{c, ld, k}, r_{s, ld, k}, f_{c, ld, k}, f_{s, ld, k}, r_{c, lu, k}, r_{s, lu, k}, f_{c, lu, k}, f_{s, lu, k}, v_{x, l, k}, v_{y, l, k} \right\} \\ 
        & \left\{ x_{k}, y_{k}, r_{c, rd, k}, r_{s, rd, k}, r_{c, ru, k}, r_{s, ru, k}, x_{r, k}, y_{r, k}, d_{r, k} \right\} \\ 
        & \left\{ r_{c, k}, r_{c, k+1}, r_{s, k}, r_{s, k+1}, f_{c, k}, f_{s, k} \right\} \\ 
        & \left\{ r_{c, rd, k}, r_{s, rd, k}, r_{c, ru, k}, r_{s, ru, k}, x_{r, k}, y_{r, k} \right\} \\ 
        & \left\{ r_{c, rd, k}, r_{s, rd, k}, f_{c, rd, k}, f_{s, rd, k}, r_{c, ru, k}, r_{s, ru, k}, f_{c, ru, k}, f_{s, ru, k}, v_{x, r, k}, v_{y, r, k}  \right\}\\ 
        & \left\{ x_{k}, x_{k+1}, y_{k}, y_{k+1}, f_{c, k}, f_{s, k}, x_{l, k}, y_{l, k}, v_{x, l, k}, v_{y, l, k}, d_{l, k}, v_{rel, l, k}, \lambda_{n, l, k}, \lambda_{t, l, k} \right\} \\ 
        & \left\{ r_{c, ld, k}, r_{c, ld, k+1}, r_{s, ld, k}, r_{s, ld, k+1}, f_{c, ld, k}, f_{s, ld, k}, r_{c, lu, k}, r_{c, lu, k+1}, r_{s, lu, k}, r_{s, lu, k+1}, f_{c, lu, k}, f_{s, lu, k} \right\} \\ 
        & \left\{ x_{k}, x_{k+1}, y_{k}, y_{k+1}, f_{c, k}, f_{s, k}, x_{l, k}, y_{l, k}, x_{r, k}, y_{r, k}, \lambda_{n, l, k}, \lambda_{t, l, k}, \lambda_{n, r, k}, \lambda_{t, r, k} \right\} \\ 
        & \left\{ x_{k}, x_{k+1}, y_{k}, y_{k+1}, f_{c, k}, f_{s, k}, x_{r, k}, y_{r, k}, v_{x, r, k}, v_{y, r, k}, d_{r, k}, v_{rel, r, k}, \lambda_{n, r, k}, \lambda_{t, r, k} \right\} \\ 
        & \left\{ r_{c, rd, k}, r_{c, rd, k+1}, r_{s, rd, k}, r_{s, rd, k+1}, f_{c, rd, k}, f_{s, rd, k}, r_{c, ru, k}, r_{c, ru, k+1}, r_{s, ru, k}, r_{s, ru, k+1}, f_{c, ru, k}, f_{s, ru, k} \right\} \\ 
        & \left\{ x_{k+1}, y_{k+1}, r_{c, ld, k+1}, r_{s, ld, k+1}, r_{c, lu, k+1}, r_{s, lu, k+1}, r_{c, rd, k+1}, r_{s, rd, k+1}, r_{c, ru, k+1}, r_{s, ru, k+1} \right\}
     \end{align}
\end{subequations}

\end{document}